\newif\ifdraft
\draftfalse
\newif\ifcameraready
\camerareadytrue

\documentclass[camera,letterpaper,nomarginnotes,nonarrowgutter]{jpaper}

\usepackage{xcolor, soul}
\usepackage{xspace}
\usepackage{multirow}
\usepackage{tabularx}
\usepackage{makecell}
\usepackage{colortbl}
\usepackage{tikz}
\usepackage[bottom]{footmisc}
\newcommand*\circled[1]{\tikz[baseline=(char.base)]{\node[shape=circle,fill,inner sep=0.5pt] (char) {\textcolor{white}{#1}};}}

\usepackage{amsmath}

\usepackage{subcaption}
\usepackage{fancyhdr}
\usepackage{mathtools}

\usepackage{booktabs}

\usepackage{pgfplots}
\pgfplotsset{compat=1.17} %

\usepackage{stfloats}

\usepackage{enumitem}

\usepackage{pifont}

\definecolor{tablegrey}{HTML}{EFEFEF}

\newcolumntype{Y}{>{\centering\arraybackslash}X}
\usepackage{tikz}

\usepackage{blindtext,graphicx}
\usepackage[absolute]{textpos}
\usepackage{datetime}

\usetikzlibrary{patterns}
\usepackage[precision=2, unit=mm]{lengthconvert}

\newcommand{\allRAG}{~\cite{zeng2024good, rackauckas2024rag, jin2024flashrag, wang2024m, li2022survey, lewis2020retrieval, gao2023retrieval, izacard2023atlas, gao2022precise, jiang2023active, fang2024ace, kim2024you, yan2024corrective, yu2024evaluation, chen2024benchmarking, cheng2024lift, caffagni2024wiki,quinn2025accelerating,sawarkar2024blended}}

\newcommand{\modernLLMs}{~\cite{dubey2024llama,liu2024deepseek,touvron2023llama}}

\newcommand{\manyANNS}{~\cite{li2019approximate, indyk1998approximate, malkov2014approximate, wang2021comprehensive, liu2004investigation, hajebi2011fast, ferhatosmanoglu2001approximate, kushilevitz1998efficient,fu2017fast,quinn2025accelerating}}

\newcommand{\manyISP}{~\cite{liang2022vstore, wang2023storage, hu2022ice, xu2023proxima}}

\newcommand{\embeddingDimensions}{~\cite{lee2024nv, wang2022text, behnamghader2024llm2vec, wang2023improving, meng2024sfrembedding, emb2024mxbai, lee2024gecko, muennighoff2024generative,muennighoff2022mteb}}

\newcommand{\manyCXL}{~\cite{jang2023cxl,li2023pond,das2024introduction,ahn2022enabling,gouk2023memory}}

\begin{document}

\title{REIS: A High-Performance and Energy-Efficient Retrieval System with In-Storage Processing}
\author
{{Kangqi Chen\textsuperscript{1}}\qquad
{Andreas Kosmas Kakolyris\textsuperscript{1}}\qquad
{Rakesh Nadig\textsuperscript{1}}\qquad
{Manos Frouzakis\textsuperscript{1}}\\
{Nika Mansouri Ghiasi\textsuperscript{1}}\qquad
{Yu Liang\textsuperscript{1}}\qquad
{Haiyu Mao\textsuperscript{1,2}}\\
{Jisung Park\textsuperscript{3}}\qquad
{Mohammad Sadrosadati\textsuperscript{1}}\qquad
{Onur Mutlu\textsuperscript{1}}
\vspace{0mm}\\\\
{ETH Z{\"u}rich\textsuperscript{1}}\qquad{King's College London\textsuperscript{2}}\qquad{POSTECH\textsuperscript{3}}
}

\maketitle

    \renewcommand{\headrulewidth}{0pt}
    \fancypagestyle{firstpage}{
        \fancyhead{} 
        \fancyhead[C]{
      } 
    \renewcommand{\footrulewidth}{0pt}
    }
  \thispagestyle{firstpage}
 
\pagenumbering{arabic}

\newcounter{version}
\setcounter{version}{999}

\begin{abstract}
Large Language Models (LLMs) face an inherent challenge: their knowledge is confined to the data that they have been trained on. This limitation, combined with the significant cost of retraining renders them incapable of providing up-to-date responses. 
To overcome these issues, Retrieval-Augmented Generation (RAG) complements the static training-derived knowledge of LLMs with an external knowledge repository.
RAG consists of three stages: (i) indexing, which creates a database that facilitates similarity search on text embeddings, (ii) retrieval, which, given a user query, searches and retrieves relevant data from the database and (iii) generation, which uses the user query and the retrieved data to generate a response. 

The retrieval stage of RAG in particular becomes a significant performance bottleneck in inference pipelines. 
In this stage, (i) a given user query is mapped to an embedding vector and (ii) an Approximate Nearest Neighbor Search (ANNS) algorithm searches for the most semantically similar embedding vectors in the database to identify relevant items.
Due to the large database sizes, ANNS incurs significant data movement overheads between the host and the storage system.
To alleviate these overheads, prior works propose In-Storage Processing (ISP) techniques that accelerate ANNS workloads by performing computations inside the storage system.
However, existing works that leverage ISP for ANNS (i) employ algorithms that are \emph{not} tailored to ISP systems, (ii) do not accelerate data retrieval operations for data selected by ANNS, and (iii) introduce significant hardware modifications to the storage system, limiting performance and hindering their adoption.

We propose \emph{REIS}, the first \underline{Re}trieval system tailored for RAG with \underline{I}n-\underline{S}torage processing that addresses the limitations of existing implementations with three key mechanisms. First, REIS employs a database layout that links database embedding vectors to their associated documents, enabling efficient retrieval. Second, it enables efficient ANNS by introducing an ISP-tailored algorithm and data placement technique that: (i) distributes embeddings across all planes of the storage system to exploit parallelism, and (ii) employs a lightweight  Flash Translation Layer (FTL) to improve performance. Third, REIS leverages an ANNS engine that uses the \emph{existing} computational resources inside the storage system, without requiring hardware modifications. The three key mechanisms form a cohesive framework that largely enhances both the performance and energy efficiency of RAG pipelines. Compared to a high-end server-grade system, REIS improves the performance (energy efficiency) of the retrieval stage by an average of $13\times$ ($55\times$). REIS offers improved performance against existing ISP-based ANNS accelerators, without introducing any hardware modifications, enabling easier adoption for RAG pipelines.

\end{abstract}

\section{Introduction}
\label{intro}

The rapid development of Large Language Models (LLMs)~\cite{dubey2024llama,liu2024deepseek,team2024gemini,jiang2024mixtral,gptj,zhang2022opt} during the past decade has led to their widespread adoption, as witnessed by the popularity of chatbots such as ChatGPT~\cite{chatGPTopenai}, Gemini~\cite{team2023gemini,team2024gemini} and DeepSeek~\cite{liu2024deepseek}. Despite this progress, modern LLMs remain limited in generating responses only from data present in their training sets. The significant cost and hardware requirements\modernLLMs{} of training further compound this problem, making frequent retraining on new data impractical, thus limiting the effectiveness of LLMs in especially domain-specific and real-time scenarios~\cite{ji2023survey, ji2023towards, martino2023knowledge}. 

Retrieval-Augmented Generation (RAG) \allRAG{}  presents a compelling solution to this problem by leveraging information retrieval techniques to feed relevant content from a document database into LLMs for text generation. At inference-time, RAG systems retrieve documents from the database that are relevant to user queries, using these to complement the training-derived knowledge of LLMs and generate contextually relevant responses.
Many recent works demonstrate the applicability of RAG to fields such as healthcare \cite{xiong2024benchmarking, unlu2024retrieval, zakka2024almanac, lozano2023clinfo}, law \cite{wiratunga2024cbr, hou2024clerc, louis2024interpretable, chouhan2024lexdrafter}, finance \cite{yepes2024financial, zhang2023enhancing, zhao2024optimizing}, and scientific research \cite{lala2023paperqa, wang2022retrieval}.

The general workflow of RAG consists of a pipeline comprised of three stages: (i) indexing, (ii) retrieval, and (iii) generation. 
First, the indexing stage is an \emph{offline} process that builds a \emph{vector database} of high-dimensional embeddings~\cite{muennighoff2022mteb, thakur2021beir, lee2024gecko, openai2024new, emb2024mxbai, embedv3, lee2024nv}. Indexing employs algorithms that cluster similar data or create graph-like structures~\cite{zobel2006inverted,coster2002inverted,malkov2018hnsw,lshDasgupta}, in order to facilitate future search operations on the data.
Second, for each incoming query, the retrieval stage identifies document chunks that are semantically relevant to the query. 
To perform this, RAG employs a process known as \textit{dense retrieval}~\cite{zhao2024dense,huang2020embedding,gan2023binary}, which encodes the query in the same vector space as document chunks and performs a similarity search between the query and the database embeddings.
Third, the generation stage feeds both the identified document chunks and the query into the LLM to generate the final response.

Although dense retrieval enables accurate semantic similarity comparison between incoming queries and document chunks~\cite{gao2022precise, thakur2023injecting, zhao2024dense, thakur2021beir}, the large embedding space results in expensive distance computations.
For RAG pipelines, a retriever that achieves \emph{both} high recall and low latency is essential because it (i) determines the quality of generated responses, and (ii) resides in the critical path of the response generation process. To strike a balance between these two conflicting objectives, RAG commonly performs dense retrieval with \textit{Approximate Nearest Neighbor Search} (ANNS) techniques, e.g., \manyANNS{}. Examples of such techniques are: (i) employing data structures that accelerate the search \cite{zobel2006inverted, coster2002inverted, zhan2022learning, malkov2018hnsw}, and (ii) quantizing data to reduce the computational complexity \cite{jegou2010product, reimers2022cohere} of search operations without significantly affecting recall.

The reduced computational complexity of ANNS renders I/O data transfers a significant bottleneck that limits search performance\manyISP. As a result, several works \manyISP{} propose In-Storage Processing (ISP) as a promising solution to accelerate ANNS-based workloads \cite{zhang2018visual, huang2020embedding, zhang2022uni, li2021embedding, gan2023binary, fu2017fast}. In particular, NDSearch~\cite{wang2023storage} demonstrates that (i) storage I/O accounts for up  to 75\% of the end-to-end ANNS latency, and (ii) ISP improves ANNS performance by 31.7$\times$ over a conventional CPU-based system, effectively mitigating the aforementioned I/O bottleneck.

We empirically make a similar observation for RAG pipelines, where the ANNS-based retrieval stage becomes the performance bottleneck due to substantial I/O overheads, as presented in Sec.~\ref{mot:io}. 
For example, when examining a RAG database containing 41.5 million document entries \cite{dataset1}, the I/O traffic from the storage system accounts for 84\% of the overall latency of the entire RAG pipeline. 
Although various software and hardware solutions that reduce the storage footprint do exist, these approaches are either unscalable (e.g., quantization methods \cite{ms_binary, gan2023binary, qdrant_binary}) or unsustainable (e.g., memory expansion \cite{jang2023cxl}). 
We conclude that In-Storage Processing (ISP) techniques are essential for fundamentally addressing the critical I/O data movement bottleneck in RAG pipelines.

Existing ISP-based ANNS accelerators~\manyISP{} face three key limitations that hinder their application to RAG workloads.
First, previous works employ search algorithms that cause performance degradation in ISP systems.
Graph-based algorithms \cite{malkov2018hnsw, jayaram2019diskann} used by ISP accelerators \cite{liang2022vstore, wang2023storage} perform searches using graph traversal, a sequential process. 
During graph traversal, the algorithm determines the next vertex to visit in the graph based on the analysis of the vertex currently being visited. However, this process exhibits irregular data~\cite{faldu2019closer,faldu2020domain} access patterns, complicating optimization and efficient execution in ISP systems.
Second, existing ISP schemes mainly focus on accelerating ANNS, the search stage in RAG applications, without optimizing the document retrieval stage, which, as we show in Sec.~\ref{mot:limitations}, contributes significant latency to the RAG pipeline. Third, in their quest to accelerate ANNS applications, existing ISP schemes introduce significant storage~\cite{hu2022ice} or hardware~\cite{mailthody2019deepstore} overheads.

\textbf{Our goal} is to fundamentally alleviate the I/O data movement bottlenecks in the retrieval stage of the RAG pipeline. To this end, we propose REIS, A \underline{Re}trieval system with \underline{I}n-\underline{S}torage Processing that employs three new key ideas: 1) an efficient ISP implementation of the clustering-based Inverted File (IVF) algorithm~\cite{zobel2006inverted, coster2002inverted, zhan2022learning} to improve end-to-end retrieval performance, 2) a new low-cost hardware-assisted mechanism in the storage system to link embeddings to their corresponding document chunks, enabling their faster retrieval, 3) a customized in-storage ANNS computation engine using the already available resources within a modern storage system to enhance the energy efficiency of the retrieval process without additional hardware.

\textbf{Key Mechanism.} 
To implement the aforementioned ideas, REIS leverages three key mechanisms.

First, we propose an ISP-tailored data placement technique and execution flow that take into account the properties of the Inverted File (IVF) algorithm~\cite{zobel2006inverted,coster2002inverted}. Since IVF organizes embeddings into clusters of similar vectors, our data placement technique (i) stores embeddings contiguously, reducing the address translation overhead from the Flash Translation Layer (FTL) and, (ii) distributes embeddings across planes to exploit the available parallelism. To execute the IVF algorithm, REIS uses: (i) the existing logic within the planes to calculate the similarity between embeddings and (ii) the SSD controller to identify the most similar embeddings.
Second, to efficiently link embeddings to documents, REIS employs a new database layout, that (i) stores embeddings and document chunks in separate regions, and (ii) creates connections between the two using the Out-Of-Band (OOB) area of the NAND Flash array, enabling efficient document retrieval.
Third, we customize the ANNS engine by using binary quantization~\cite{gan2023binary,ms_binary,shakir2024quantization} and a hybrid SSD design~\cite{chang2010hybrid}. Binary quantization reduces the computational complexity of ANNS, while the hybrid SSD design combines reliable ISP with high storage density. Specifically, our hybrid SSD design employs (i) SLC, using Enhanced SLC programming~\cite{park2022flash} for high-performance and reliable In-Storage computation on embeddings and (ii) TLC for storing document chunks at high density.

\textbf{Key Results.} We evaluate REIS on two SSD configurations based on a cost-~\cite{pm9a3} and a performance-oriented~\cite{micron9400} SSD design. 
We compare its performance and energy efficiency against a high-end 256-core CPU system on two commonly used benchmark datasets from BEIR~\cite{thakur2021beir} and a large-scale public dataset \cite{dataset1}, demonstrating that REIS (i) achieves an average speedup of $13\times$ and up to $112\times$, and (ii) improves energy efficiency by an average of $55\times$ and up to $157\times$.  Compared to a state-of-the-art ISP-based ANNS accelerator~\cite{hu2022ice} REIS yields an average speedup of $21.4\times$ ($7.67\times$) and $24.2\times$ ($9.76\times$) at $0.98$ ($0.90$) \(Recall@10\) across all evaluated datasets for the cost- and performance-oriented SSDs, respectively.
Since REIS does not introduce any hardware modifications to the storage system, its adoption for RAG is much easier than prior ISP-based accelerators.

The contributions of this work are listed as follows:
\begin{itemize}
\item This is the first work to quantitatively evaluate the large performance overheads of I/O data movement in the retrieval stage of the Retrieval-Augmented Generation pipeline.
\item We comprehensively analyze the limitations of existing techniques that aim to alleviate the I/O data movement bottleneck of the RAG pipeline. 
We identify two major issues that make integrating existing ISP-based ANNS accelerators into the RAG pipeline inefficient and impractical.
\item We propose REIS, the first ISP-based retrieval system tailored for RAG. REIS (i) supports efficient document retrieval by building the correlation between embeddings and documents within the storage system, (ii) improves retrieval performance by introducing an ISP-friendly algorithm, and (iii) improves energy and area efficiency via a customized in-storage ANNS computation engine using computational resources already available in a modern storage system.

\item We implement REIS based on a cost- and a performance-oriented SSD design and evaluate its performance and energy efficiency. Against a 256-core CPU system, REIS provides an average speedup (energy efficiency improvement) of $13\times$ ($55\times$). Compared to a state-of-the-art ISP-based ANNS accelerator, REIS accelerates RAG retrieval from $7.67\times$ and up to $24.1\times$ depending on (i) the SSD configuration used, and (ii) the target \(Recall@10\) value.

\end{itemize}

\section{Background}

\subsection{Retrieval-Augmented Generation}
\label{bg:rag}

Retrieval-Augmented Generation (RAG)\allRAG{} is the process of incorporating knowledge from an external document database into LLM inference. To identify the most relevant data, RAG employs dense retrieval~\cite{zhao2024dense,gao2022precise}, %
a similarity search operation on dense vectors representing the semantics of the text, called embeddings~\cite{lee2024nv, wang2022text, behnamghader2024llm2vec, wang2023improving, meng2024sfrembedding, emb2024mxbai, lee2024gecko, muennighoff2024generative}. %
To encode this information, embeddings feature high dimensionality, often containing 768 to 8192 dimensions\embeddingDimensions{}.

As mentioned in Sec.~\ref{intro}, RAG is a pipeline comprised of three stages: (i) indexing, (ii) retrieval, and (iii) generation. 
Indexing creates data structures such as clusters or graphs, that facilitate faster semantic similarity search on the embeddings~\cite{zobel2006inverted,coster2002inverted,malkov2018hnsw,lshDasgupta}. 
In the retrieval step, the RAG system receives a query and encodes it as an embedding. It then searches the database for the $k$ most similar embeddings,
with $k$ being a parameter specified by the system. Once the most similar embeddings are identified, the RAG system retrieves the corresponding document chunks. In the generation stage, both the retrieved document chunks and the query are fed to the LLM in order to perform inference and generate a response.

While the main application for RAG currently is document retrieval for question answering~\cite{chen2020open}, researchers have also proposed multi-modal RAG pipelines~\cite{zhong2022training, wang2024m, he2024multimodal}. For example, %
Vision Transformers~\cite{dosovitskiy2020image} %
enable joint image and text retrieval~\cite{radford2021learning, jia2021scaling, zhou2024vista}. 
Other works~\cite{girdhar2023imagebind} combine even more modalities into the same embedding space such as audio, depth, thermal and movement data.

\subsection{Approximate Nearest Neighbor Search}
\label{bg:anns}

The retrieval stage forms a critical bottleneck in the RAG pipeline, as generation cannot begin before the relevant document chunks have been retrieved. %
The simplest method of identifying the $k$ most relevant (top-$k$) embeddings is \textit{Nearest Neighbor Search} (NNS), which entails: (i) calculating the distances (e.g. Euclidean Distance~\cite{anastasiu2015l2knng,arya1998optimal}) between the query and all database embeddings and (ii) selecting the $k$ database embeddings %
with the lowest distance. However, %
a brute-force approach incurs significant computational overheads due to (i)~the large size of embedding vectors~\cite{muennighoff2022mteb} and (ii)~the large number of database embeddings, reaching multiple millions~\cite{yang2018hotpotqa,dataset1} and even billions~\cite{billionScaleRag,jansen2022perplexed}, resulting in expensive distance computations. To accelerate the retrieval stage, RAG often performs \textit{Approximate Nearest Neighbor Search} (ANNS)~\cite{li2019approximate}, trading off some retrieval accuracy for faster similarity search. 
To quantify this drop in accuracy, researchers often use the \(Recall@k\) metric~\cite{chen2022approximate, manohar2024parlayann, wang2021comprehensive, liu2004investigation, li2019approximate}, which is defined the %
fraction of how many of the $k$ most relevant document chunks have actually been retrieved by ANNS.

Two popular methods for performing ANNS are (i) quantization~\cite{jegou2010product,reimers2022cohere,gan2023binary,ms_binary,qdrant_binary} and (ii) algorithm-based techniques\manyANNS{}. Quantization methods compress data, reducing their storage footprint and speeding up computation. For Example, Product Quantization (PQ)~\cite{jegou2010product} partitions large embedding vectors into smaller sub-vectors and assigns each sub-vector to a cluster. PQ then concatenates the IDs of the clusters into a new vector that represents the original vector. Binary Quantization (BQ) compresses each embedding component from its original floating-point precision (e.g., FP32) down to a single bit, achieving a $32\times$ compression ratio. Recent studies~\cite{muennighoff2022mteb, reimers2022cohere, shakir2024quantization, qdrant_binary} show that BQ accelerates ANNS by up to $40\times$, with a small impact on recall when combined with a low-cost rescoring step~\cite{reimers2022cohere}.

Algorithm-based methods organize data by clustering them or creating graph-like data structures, which can be searched efficiently without traversing the entire database. For example, the Inverted-File (IVF) algorithm~\cite{zobel2006inverted, coster2002inverted, zhan2022learning} organizes embeddings into clusters that are each represented by a centroid. To perform a search for a given query embedding, first a \textit{coarse-grained} search identifies the cluster centroids closest to the query embedding. Second, a \textit{fine-grained} search on all embeddings within these clusters (approximately) yields the closest neighbors to the query embeddings. Other algorithms also exist, such as: (i) \textit{Hierarchically Navigable Small World} (HNSW)~\cite{malkov2018hnsw}, which constructs a hierarchy of graphs, where higher and lower levels of the hierarchy direct the search in a coarse- and fine-grained manner, respectively, and (ii) Locality-Sensitive Hashing (LSH)~\cite{lshDasgupta}, which hashes similar embeddings into the same bucket with high probability.

\subsection{SSDs \& NAND Flash Memory}

\begin{figure}[htbp]
    \captionsetup{aboveskip=5pt, belowskip=0pt}
    \centering
    \includegraphics[width=\linewidth]{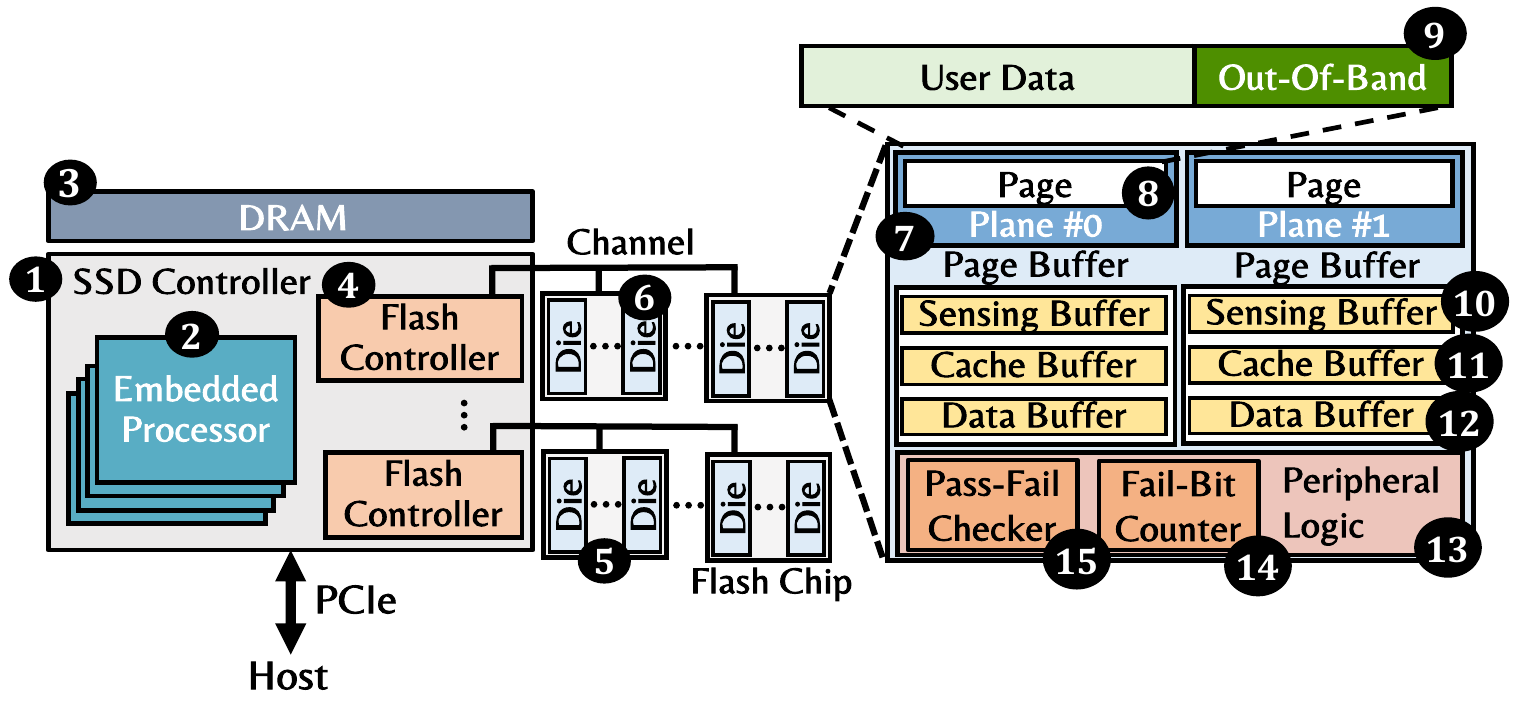}
    \caption{NAND Flash Memory Architecture}
    \label{fig:bg1}
\end{figure}

Figure~\ref{fig:bg1} presents an overview of a modern SSD architecture based on NAND flash memory.
An SSD comprises of an SSD controller, DRAM and multiple NAND flash chips.
The SSD controller (\circled{1})~\cite{cai-procieee-2017, cai2017vulnerabilities, cai-iccd-2012, cai-errors-2018, cai-insidessd-2018, agrawal2008design, tavakkol2018mqsim} handles the I/O requests from the host, and performs maintenance tasks such as garbage collection (e.g.,~\cite{yang2014garbage, cai2017error, tavakkol2018mqsim, agrawal2008design, shahidi2016exploring, lee2013preemptible,jung2012taking, choi2018parallelizing,wu2016gcar,cai-errors-2018,cai2017vulnerabilities}) and wear-leveling (e.g.,~\cite{shin2009ftl,lim2010faster,zhou2015efficient, agrawal2008design}). The SSD controller contains multiple embedded microprocessors (\circled{2})~\cite{arm} that execute the firmware called the Flash Translation Layer (FTL)~\cite{gupta2009dftl, tavakkol2018mqsim, lim2010faster, shin2009ftl, zhou2015efficient}. The SSD controller stores metadata (e.g., logical-to-physical page mapping table~\cite{gupta2009dftl}) and frequently-accessed pages in %
a DRAM (\circled{3}) internal to the SSD.
The DRAM size is typically 0.1\% of the storage capacity (e.g., 1GB DRAM for each TB of storage capacity \cite{samsung990pro}).
The SSD controller translates the logical page address of each I/O request to a physical page address, and issues commands to the flash chips~\cite{cai2017error,agrawal2008design,nadig2023venice} via the flash controllers.
An SSD consists of multiple flash controllers (\circled{4})~\cite{kim2021decoupled, kim2022networked,wu2012reducing, micheloni2010inside, microchip-16-channel-controller}, which are embedded processors that interface the SSD controller with flash chips (\circled{5}). Each flash controller is responsible for communication with multiple flash chips sharing the same channel.
The flash controller selects a flash chip for read/write operations and initiates command and data transfers.

Each NAND flash chip is comprised of multiple flash dies (\circled{6}), which operate independently of each other. 
Each die consists of 2\text{--}16 planes (\circled{7})~\cite{kim2022a, cho2022a, kouchi2020128gb} , that can perform read or write operations in parallel.
Planes are further divided into groups of blocks, with each block 
consisting of hundreds of pages (\circled{8}). 
A flash page, typically sized at 16KB, 
consists of thousands of NAND flash cells placed horizontally. A flash page typically stores user data, and consists of a dedicated out-of-band area (\circled{9})~\cite{yao2021hdftl,huang2015unified,sun2023leaftl} (e.g., 64-256 bytes) to store metadata related to error correction codes and logical-to-physical mapping.
NAND flash memory executes read and program operations at the page granularity, and performs erase operations at block granularity~\cite{liu2022flash, compagnoni2017reviewing, micheloni2010inside,cai-procieee-2017,cai-date-2012,cai-iccd-2012,park2022flash}.
A flash die employs a page buffer which acts as an intermediate buffer during read and write operations. 
The page buffer consists of multiple buffers~\cite{leong-uspatent-2008, macronix-technote-2013, micron-datasheet-2009, samsung-datasheet-2009, toshiba-datasheet-2012, micheloni-insidenand-2010, higuchi202130, khakifirooz202130,shibata20191} (e.g., three buffers if each flash cell stores 3 bits) to store the bits in a flash page. 
The sensing buffer (\circled{10}) is the main buffer that temporarily stores flash page data during the read operation. 
The cache buffer (\circled{11}) improves read performance by enabling data transfer from the flash chip to the flash controller in parallel with the next read operation. 
Data buffers (\circled{12}) are typically used when (1) programming multiple bits per cell, and (2) reading a single bit from a flash cell that stores multiple bits.

Based on the number of bits stored in a flash cell, a flash cell can be classified as a single-level cell (SLC; 1 bit)~\cite{cho2001dual}, multi-level cell (MLC; 2 bits)~\cite{lee20043}, triple-level cell (TLC; 3 bits)~\cite{maejima2018512gb, kim2022a, cho2022a}, or quad-level cell (QLC; 4 bits)~\cite{cho20221}. 
While the SSD capacity increases as each flash cell stores more bits, the increased value density leads to higher latency and lower endurance~\cite{cai2017vulnerabilities, cai2017error, micheloni2010inside, cai2012error, cai-errors-2018}.
To enable reliable writes to flash cells, SSD manufacturers use Incremental Step Pulse/Erasure Programming (ISPP/ISPE) techniques~\cite{suh-ieeejssc-1995, jeong-fast-2014}. ISPP/ISPE technique iterates through multiple steps of gradually inserting/ejecting electrons into/from the flash cell until the desired charge level is reached. 
The peripheral circuitry (\circled{13}) in each flash die includes an on-chip digital bit counter,  pass/fail checking logic and XOR logic between the latches.
The digital bit counter and pass/fail checking logic~\cite{cho2024aero, micheloni2010inside, choi2020modeling} are used to test the state of the cells and guide the ISPP/ISPE process. 
To further increase reliability, the XOR logic between the latches is used for on-chip data randomization~\cite{park2022flash, kim201221, hu2022ice, lu2018fault, cao2022efficient}.

\subsection{In-Storage Processing}

In-storage processing (ISP) is a computation paradigm that enables processing of data within the storage device.
ISP techniques provide significant performance and energy efficiency benefits over conventional systems for data-intensive applications, such as genomics~\cite{mansouri2022genstore, ghiasi2024megis}, neural networks~\cite{kim2023optimstore, jang2024smart, wang2024beacongnn, niu2024flashgnn}, databases~\cite{kim2016storage, gu2016biscuit, jo2016yoursql, chan-signmod-1998, oneil-ideas-2007, li-vldb-2014, li-sigmod-2013, goodwin-SIGIR-2017, seshadri-micro-2013, seshadri-micro-2017, seshadri-ieeecal-2015, hajinazar-asplos-2021, FastBitA9, wu-icde-1998, guz-ndp-2014,redis-bitmaps} and graph analytics~\cite{perach-arxiv-2022, seshadri-micro-2017, jun-isca-2015, torabzadehkashi-pdp-2019, lee-ieeecal-2020, beamer-SC-2012, besta-micro-2021, li-dac-2016, gao-micro-2021, hajinazar-asplos-2021}. 
Unlike conventional systems, ISP techniques leverage the high internal bandwidth of the storage system and reduce data movement across the memory hierarchy.
ISP techniques perform computation by (1) leveraging the embedded general-purpose cores (e.g.,~\cite{seshadri-osdi-2014,mansouri-asplos-2022,kang-msst-2013,gu-isca-2016,wang-eurosys-2019,acharya-asplos-1998,keeton-sigmod-1998, wang-damon-2016, koo-micro-2017, tiwari-fast-2013, tiwari-hotpower-2012, boboila-msst-2012, bae-cikm-2013,torabzadehkashi-ipdpsw-2018, kang-micro-2021,li-atc-2021,lim-icce-2021,kim-sigops-2020}) already present in the SSDs, or (2) placing hardware accelerators (e.g., ~\cite{jun-isca-2018, mailthody-micro-2019,pei-tos-2019, do-sigmod-2013,kim-infosci-2016, riedel-computer-2001,riedel-vldb-1998, torabzadehkashi-pdp-2019,liang-atc-2019,cho-wondp-2013, jun-isca-2015, lee-ieeecal-2020, ajdari-hpca-2019,liang-fpl-2019,jeong-tpds-2019,jun-hpec-2016}) near the flash chips.

Prior works (e.g.,~\cite{seshadri-osdi-2014,mansouri-asplos-2022,kang-msst-2013,gu-isca-2016,wang-eurosys-2019,acharya-asplos-1998,keeton-sigmod-1998, wang-damon-2016, koo-micro-2017, tiwari-fast-2013, tiwari-hotpower-2012, boboila-msst-2012, bae-cikm-2013,torabzadehkashi-ipdpsw-2018, kang-micro-2021,li-atc-2021,lim-icce-2021,kim-sigops-2020}) propose techniques to utilize the embedded cores in the SSD for computations such as filtering, aggregation, and encryption. These general-purpose embedded cores are beneficial only for simple computations because the primary responsibility of these cores is to execute the FTL and handle I/O requests.
A large body of prior work (e.g.,~\cite{jun-isca-2018, mailthody-micro-2019,pei-tos-2019, do-sigmod-2013,kim-infosci-2016, riedel-computer-2001,riedel-vldb-1998, torabzadehkashi-pdp-2019,liang-atc-2019,cho-wondp-2013, jun-isca-2015, lee-ieeecal-2020, ajdari-hpca-2019,liang-fpl-2019,jeong-tpds-2019,jun-hpec-2016}) proposes to embed hardware accelerators near the flash packages to accelerate application-specific computations. These hardware accelerators provide significant performance benefits, but add area and power overheads to the SSD.
Several prior works~\cite{liang2022vstore, xu2023proxima, wang2023storage, hu2022ice, mailthody2019deepstore, kim2022accelerating} identify the substantial I/O traffic (up to 70-75\% of the end-to-end search latency \cite{liang2022vstore, wang2023storage}) in billion-scale ANNS applications \cite{jayaram2019diskann, johnson2019billion, sift2010anns,simhadri2022results, bigannbenchmarks} and propose offloading ANNS to the storage system.
These  ISP-based ANNS accelerators can largely alleviate the I/O overhead, %
improving performance over conventional systems.

\section{Motivation}

A key limitation of modern LLMs is their inability to generate responses with information beyond their training data. To solve this issue, modern LLM application frameworks~\cite{LlamaIndex, LangChain, Haystack} support Retrieval-Augmented Generation, combining the text generation capabilities of LLMs~\cite{dubey2024llama,liu2024deepseek,team2024gemini,jiang2024mixtral,gptj,zhang2022opt} with an external knowledge database, as described in Sec.~\ref{bg:rag}.
Beyond information retrieval, RAG also enables long-tail knowledge memorization~\cite{mallen2022not}, alleviating the need for large models with billions of parameters \cite{izacard2023atlas}, and mitigating the risk of revealing training data \cite{zeng2024good}. 

While current research on RAG~\cite{gao2023retrieval, li2022survey, zhao2024retrieval, yu2024evaluation, chen2024benchmarking} mainly focuses on further enhancing these capabilities, to our knowledge, no existing works attempt to characterize and address the inefficiencies found in RAG pipelines. In this section, we analyze the performance bottlenecks of RAG and discuss the issues existing systems face when tackling these bottlenecks.

\subsection{Performance Bottleneck of RAG Pipelines}
\label{mot:io}

As described in Sec.~\ref{bg:rag}, RAG pipelines consist of one offline stage, indexing, and two online stages, i.e., retrieval and generation. The online stages of RAG entail (i) encoding the query into an embedding vector, (ii) performing dense retrieval for relevant documents, and (iii) using the retrieved documents and the query to generate a response. For these steps, respectively, the RAG system has to load (i) the embedding model, (ii) the RAG database and (iii) the generation model (i.e., the LLM) from the storage system. With the aim of identifying potential inefficiencies, we %
measure the latency contributions of the above stages to the RAG pipeline. 

\noindent\textbf{Methodology.} For encoding and generation we chose two popular open-source models, \textit{all-roberta-large-v1}~\cite{reimers-2019-sentence-bert, reimers-2020-multilingual-sentence-bert} and Llama 3.2 1B~\cite{dubey2024llama}, respectively.
We use FAISS~\cite{douze2024faiss} flat indexes to link embeddings to document chunks.
We evaluate RAG performance on two datasets, \textit{HotpotQA}~\cite{yang2018hotpotqa} with 5.3 million entries and the English subset of a Wikipedia-based dataset (\textit{wiki\_en})~\cite{dataset1}, with 41.5 million entries. For each query, we retrieve the 10 most relevant document chunks (top-10 retrieval). Our RAG system consists of a high-end NVIDIA A100 GPU~\cite{choquette2021nvidia} for embedding and generation and two high-end Intel Xeon Gold 5118 CPUs~\cite{intelxeon}, with a Samsung PM9A3 PCIe 4.0 SSD \cite{pm9a3} for retrieval. The system is also equipped with 1.5TB of DDR4 memory~\cite{micronddr4}. 

\begin{figure}[h]
    \centering
    \includegraphics[width=\linewidth]{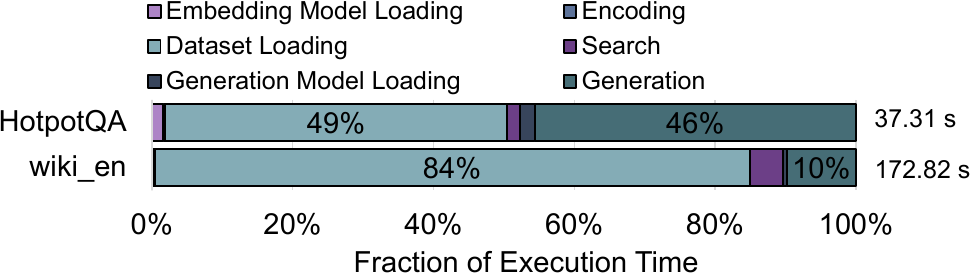}        
    \caption{Latency breakdown for a typical RAG pipeline. Total time is displayed next to each bar.}
    \label{fig:mot1}
\end{figure}

\noindent\textbf{Results.} Figure~\ref{fig:mot1} shows the contribution of different operations in the RAG pipeline to end-to-end execution time. 
We make two key observations. First, dataset loading accounts for a substantial portion of the pipeline's overall latency, reaching 84\% for \textit{wiki\_en}. 
Second, the latency attributed to dataset loading increases with dataset size. For example, as the dataset size grows by approximately 8$\times$ from \textit{HotpotQA} to \textit{wiki\_en}, the percentage of latency attributed to dataset loading increases by around 1.7$\times$. We conclude that dataset loading during the retrieval stage contributes significant latency to the RAG pipeline and becomes a performance bottleneck, especially for large datasets. We refer to this bottleneck
as the \emph{I/O data movement bottleneck} in RAG pipelines.

\begin{figure}[h]
    \centering
    \includegraphics[width=\linewidth]{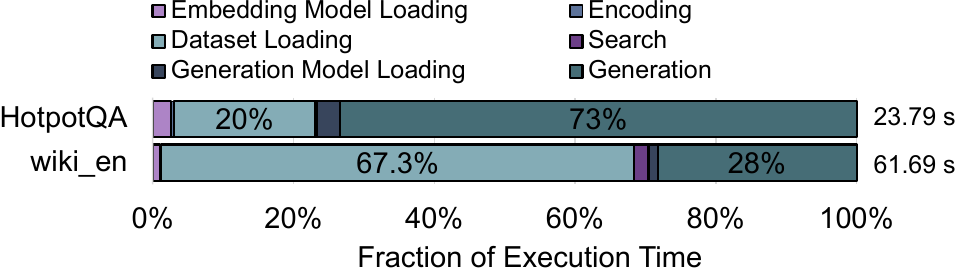}        
    
    \caption{Latency breakdown for a RAG pipeline using Binary Quantization (BQ). Total time is displayed next to each bar.}
    \label{fig:mot2}
\end{figure}

As an important caveat, we acknowledge that the contribution of \emph{I/O data movement} to end-to-end RAG performance largely depends on encoding and generation model sizes. Larger models (e.g., Llama 3.2 90B~\cite{dubey2024llama}) increase 
generation latency, due to the increased computation cost, potentially reducing the impact of I/O data movement in the RAG retrieval stage. Even in this case, I/O data movement can still bottleneck the RAG pipeline for two key reasons. First, LLM acceleration techniques~\cite{yu2022orca, rasley2020deepspeed, holmes2024deepspeed, ainslie2023gqa, dao2022flashattention, dao2023flashattention, shah2024flashattention, kwon2023vllm,wang2022tesseract,shoeybi2019megatron} and more powerful hardware~\cite{park2024attacc,nvidiaH100,nvidiaB100,sun2025lincoln} can substantially reduce generation latency, exacerbating the I/O bottleneck of the RAG pipeline. For instance, tensor parallelism \cite{kwon2023vllm, shoeybi2019megatron, wang2022tesseract} enables efficient LLM generation on multi-node GPU systems~\cite{choquette2020nvidia, tirumala2024nvidia, nvidiaH100,nvidiaB100}, significantly improving performance. Second, the increasingly popular Mixture-of-Experts (MoE) LLM architecture~\cite{liu2024deepseek,jiang2024mixtral,zhou2022mixture}
can reduce computational cost and increase generation performance of large LLMs. As a result, we anticipate that the retrieval, and not the generation stage, will remain a significant bottleneck in future RAG pipelines.

\subsection{Limitations of Existing RAG Optimizations}
\label{mot:limitations}

We discuss the limitations of existing optimizations when trying to alleviate the I/O data movement bottleneck in RAG pipelines.  

\noindent\textbf{Batching.} One possible solution is to batch multiple queries before performing retrieval to amortize dataset loading overheads. However, the effectiveness of this technique remains limited in practice as queries from different domains (e.g., medical, law, finance) must be served from \emph{different}, domain-specific~\cite{xiong2024benchmarking, unlu2024retrieval, zakka2024almanac, wiratunga2024cbr, hou2024clerc, louis2024interpretable, yepes2024financial, zhang2023enhancing, zhao2024optimizing, lala2023paperqa, wang2022retrieval} or multi-modal datasets~\cite{mazumder2021avgzslnet, xue2024retrieval, ghosh2024recap, gg2024multi, ms2024multi, pal2020pinnersage, ashutosh2023hiervl, yang2023vid2seq, zolfaghari2021crossclr} to enhance generation quality. %

\noindent\textbf{Quantization.} Quantization techniques, such as Product Quantization (PQ) or Binary Quantization (BQ), can reduce the memory footprint of RAG applications.
Recent studies~\cite{muennighoff2022mteb, reimers2022cohere, shakir2024quantization, qdrant_binary, ms_binary} demonstrate that BQ provides a good trade-off between storage footprint and recall. To further evaluate this trade-off, we repeat the %
previous experiment using BQ for the embeddings.
As shown in Fig~\ref{fig:mot2}, while BQ reduces the I/O data movement overhead by 17-29\% for both datasets, dataset loading remains the bottleneck for the larger \textit{wiki\_en} dataset, amounting to 67\% of the total latency.

While quantization significantly reduces the size of embeddings, this is not possible for the document chunks, which amount to 9GB of the total 14GB transferred for the \textit{wiki\_en} dataset (after BQ on the embeddings). Therefore, we conclude that quantization techniques are useful in reducing the I/O data movement bottleneck, but they cannot eliminate it.

\noindent\textbf{Algorithmic Optimization.} ANNS algorithms often improve retrieval performance by using sophisticated indexes~\cite{zobel2006inverted,coster2002inverted,malkov2018hnsw,lshDasgupta}, which reduces search time. The data structures used to store these indexes are often larger than the flat indexes used for simple brute-force approaches, potentially exacerbating the I/O data movement bottleneck.
Hybrid ANNS algorithms~\cite{jayaram2019diskann,chen2021spann} attempt to overcome the I/O data movement bottleneck by storing the index in SSDs and loading parts of it in memory for distance computations on demand. SPANN~\cite{chen2021spann} provides the state-of-the-art performance-accuracy tradeoff among hybrid ANNS solutions, enabling small amounts of DRAM (e.g. 32GB) to accelerate searches in TB-sized SSD-resident datasets. Specifically, SPANN groups embeddings into clusters and stores them in the SSD, only keeping cluster centroids in memory. 
We conduct an experimental study on SPANN and find two major limitations of this type of solution.
First, we  observe that achieving a reasonable recall-accuracy tradeoff requires selecting a large number of centroids, increasing memory footprint and lowering performance. For example, reaching $0.92$ $Recall@10$ in \textit{HotpotQA} requires storing 24\% of all embeddings as centroids in memory, yielding only a 22\% speedup over exhaustive search. This observation also matches with the original study of this algorithm~\cite{chen2021spann}.  
Second, hybrid ANNS algorithms such as SPANN only optimize storage and retrieval for embeddings and not for the document chunks of a vector database.  
We conclude that hybrid ANNS algorithms also do \emph{not} fundamentally alleviate the I/O data movement bottleneck.

\noindent\textbf{Memory Expansion.} As our analysis in Sec.~\ref{mot:io} shows, data movement between storage and the host contributes significant latency to the RAG retrieval stage. Memory expansion techniques such as those enabled by Compute Express Link (CXL)\manyCXL{} enable very large memory capacities that could theoretically keep RAG datasets resident in memory. However, such approaches suffer from two key drawbacks. First, main memory is significantly (i.e., more than an order of magnitude) more expensive per GB than flash storage, at approximately 3.10~\cite{samsung128GBDDR4} vs 0.1~\cite{pm9a3} USD per GB, respectively. Second, such approaches are unsustainable as (i) continuously increasing dataset sizes, and (ii) the growing number of datasets for domain-specific applications~\cite{xiong2024benchmarking, unlu2024retrieval, zakka2024almanac, wiratunga2024cbr, hou2024clerc, louis2024interpretable, yepes2024financial, zhang2023enhancing, zhao2024optimizing, lala2023paperqa, wang2022retrieval} eventually overwhelm the capacity of such systems.

\noindent\textbf{ANNS Acceleration Inside the Storage.}
Prior works propose In-Storage processing (ISP) techniques~\cite{mailthody2019deepstore, liang2022vstore, wang2023storage} to alleviate the I/O data movement bottleneck in the ANNS kernel. Although ANNS forms a key component of RAG, existing ISP-based ANNS accelerators cannot entirely eliminate the I/O data movement bottleneck for three key reasons.
First, prior ANNS acceleration works~\cite{mailthody2019deepstore, liang2022vstore, wang2023storage} employ graph-based algorithms such as HNSW~\cite{malkov2018hnsw} and DiskANN~\cite{jayaram2019diskann}, using graph-traversal to identify similar neighbors.
During graph traversal, the algorithm performs an analysis on the current vertex to identify the next vertex. As a result, graph traversal induces irregular access patterns~\cite{faldu2019closer,faldu2020domain} that underutilize the internal bandwidth of the SSD due to costly channel and NAND Flash chip conflicts~\cite{nadig2023venice,kim2022networked}. Second, prior ISP-based ANNS accelerators~\cite{hu2022ice, mailthody2019deepstore, liang2022vstore, wang2023storage} focus primarily on accelerating the search operation without providing efficient support for retrieving the associated documents. However, as shown in Figs.~\ref{fig:mot1} and~\ref{fig:mot2}, the dataset loading step contributes significant latency to RAG retrieval.
Third, works such as~\cite{hu2022ice,mailthody2019deepstore} introduce significant overheads storage and hardware overheads. For example, ICE~\cite{hu2022ice} in order to perform computations inside NAND flash dies, stores data in a format that can tolerate errors without error correction. This format incurs a $32\times$ ($8\times$) storage overhead for data in 8-bit (4-bit) precision, resulting in high storage overheads. Another example is DeepStore~\cite{mailthody2019deepstore}, which incurs significant area and power overheads by introducing a systolic array-based architecture in the storage system to perform query matching by executing Deep Neural Networks. Overall, these limitations hinder the adoption on ISP-based acceleration techniques in RAG pipelines.

\subsection{Our Goal}
 Based on our observations and analyses in Sec.~\ref{mot:io} and \ref{mot:limitations}, we conclude that (1) the I/O data movement of RAG significantly bottlenecks its performance, and (2) none of the prior techniques effectively eliminate this bottleneck in the RAG pipeline. 
\textbf{Our goal} is to fundamentally alleviate the I/O data movement bottleneck in RAG %
through an ISP design that does not introduce modifications to the hardware of the storage system.

\section{REIS}
\label{reis:overview}

REIS is an In-Storage Processing (ISP)-based retrieval system that alleviates the I/O data movement bottleneck in the RAG pipeline. REIS works by receiving query embeddings from the host, querying the database inside the storage, and then returning relevant document chunks, greatly reducing communication between host and storage system. 

ISP introduces two significant design challenges. First, the available embedded cores are limited in terms of both performance and functionality (e.g., lack of floating point support~\cite{armr8}). Second, the flash channel bandwidth is limited compared to the total NAND flash read bandwidth. As 
described in Sec.~\ref{reis:isse}, REIS uses the existing hardware inside the NAND flash planes to alleviate the load on the embedded cores, which, however, introduces new limitations: (I) The logic inside flash dies only supports simple bitwise and bit-counting operations. (II) NAND flash reads are unreliable, requiring the use of error correction codes (ECC)~\cite{cai2012error} to achieve robust operation. Since ECC is typically performed by the controller~\cite{arm}, performing computation inside flash dies requires fundamentally different error mitigation mechanisms.

\begin{figure}[h]
    \centering
    \includegraphics[width=\linewidth]{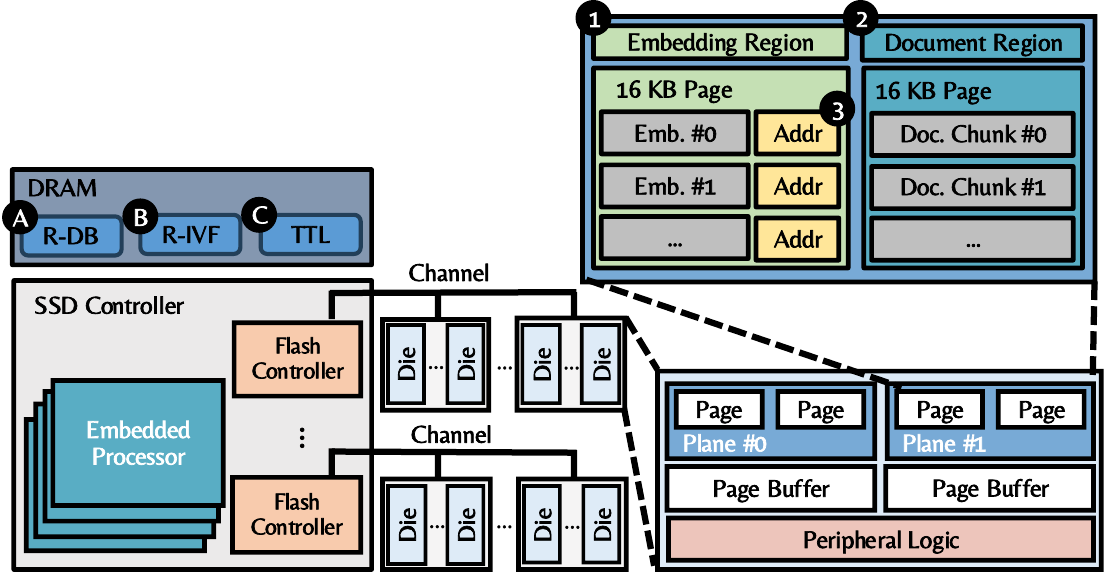}
        
    \caption{Overview of REIS.}
    \label{fig:ov}
\end{figure}

In this section, we explain the design decisions behind REIS, which alleviate the aforementioned issues. Figure~\ref{fig:ov} presents an overview of the system and the key mechanisms it consists of. First, REIS employs a vector database layout that links embeddings with documents in order to enable efficient document retrieval (Sec.~\ref{reis:dd}). Second, REIS introduces support for the Inverted File (IVF) algorithm in ISP systems, improving the end-to-end retrieval performance (Sec.~\ref{reis:ivf}). Third, an in-storage ANNS engine efficiently executes the ANNS kernel (Sec.~\ref{reis:isse}).

\subsection{Database Layout}
\label{reis:dd}
REIS introduces a vector database layout that distributes and links
embeddings and documents in order to maximize the data access parallelism for in-storage computation.
The database layout (i) distributes the vector database into an index region and a document region,
(ii) creates low-overhead links between each embedding and its associated document chunk, and (iii) provides coarse-grained access to each dataset to avoid frequent FTL invocations.

\subsubsection{Database Distribution}
\label{sub:database_distr}
During the retrieval stage of RAG, the ANNS kernel performs distance calculations on the database embeddings to select the top-$k$ most similar documents. As a result, accesses to embeddings are far more frequent than accesses to documents. Based on this observation, we distribute the database in three ways to improve the efficiency of accessing embeddings. First, we map embeddings and documents to two
separate regions of the NAND flash array, the \textit{embedding} (\circled{1} in Fig.~\ref{fig:ov}) and the \textit{document} (\circled{2}) regions, respectively.
Second, we employ Parallelism-First Page Allocation~\cite{zhang2019spa} to evenly distribute embeddings across all planes of the storage system. Third, we assign each document chunk to an individual 4KB sub-page or a 16KB page, adapting to different document chunking granularities~\cite{li2023towards, voyage2024, SFR-embedding-2, lee2024nv, muennighoff2024generative, wang2023improving, behnamghader2024llm2vec}.

\subsubsection{Hybrid SSD design}
Modern SSDs employ Triple-Level Cells (TLC) which rely on ECC to combine high density with data integrity, requiring data transfers to the embedded cores of the SSD controller for error correction. As will be shown in Sec.~\ref{reis:isse}, REIS performs operations within the planes and dies of the storage system. Thus, performing ECC on the controller would create significant data movement overheads, negating potential speedups. In order to: (i) eliminate such overheads and (ii) allow error-free in-plane embedding distance calculation without ECC, REIS employs \emph{HybridSSD}~\cite{zhang2019spa, turbowrite, tripathy2022ssd, xiao2022pasm} techniques in the ANNS engine. Specifically, we employ soft partitioning to create (i) a robust, non-ECC Single Level Cell (SLC) partition for storing binary embeddings, and (ii) a typical, high-density TLC partition that stores the database's document chunks and embeddings that are not processed within the planes (e.g INT8 embeddings for reranking). To further improve the robustness of the SLC partition, REIS makes use of the Enhanced SLC-mode Programming (ESP)~\cite{park2022flash}, which maximizes the margin between the voltage ranges of the values in SLC, achieving \emph{zero} BER \emph{without} ECC. As an added benefit, SLC programming slightly enhances RAG performance due to decreased read latency of SLC compared to TLC~\cite{turbowrite}.

\subsubsection{Embedding-Document Linkage}
\label{sub:embedding_doc_link}

While the database layout of Sec.~\ref{sub:database_distr} can increase performance by separating the frequently accessed embeddings from the less frequently accessed document chunks, performing document retrieval requires a connection between the two.
To achieve this, REIS employs a low-cost linkage mechanism within the storage system that associates each embedding with the address of its corresponding document chunk. 

Modern NAND flash memory provisions some storage space for ECC bits known as the Out-Of-Band (OOB) area (e.g., 2208 spare bytes for each 16KB page~\cite{qin2021better,raquibuzzaman2022intrablock}). During each page read, the page buffer loads OOB data together with the page. We re-purpose a small portion of the OOB area to store the address of the document chunk that is associated with each embedding (\circled{3} in Fig.~\ref{fig:ov}).
For example, assuming a dataset where (i) each embedding and document chunk occupies 4KB (i.e., a sub-page \cite{lee2022mqsim}) and  (ii) each document chunk requires a 4-byte address, linking embeddings to documents requires 16 spare bytes (or 0.7\% of the OOB area) for each page. 
This approach ensures that whenever an embedding is loaded to the page buffer, the address of its associated document chunk is also loaded.  Therefore, when the storage system conducts distance computation for a page of embeddings using the mechanisms proposed in Sec.~\ref{reis:isse}, the addresses of associated document chunks %
are available in the page buffer for document identification and retrieval.
Our proposed mechanism eliminates the need to maintain a specialized data structure for document retrieval with minimal space overhead to the storage system.

\subsubsection{Coarse-Grained Access}
\label{reis:coarsega}
With the aim of (i) distinguishing between different RAG datasets in the storage system and (ii) reducing the frequent address translation overheads when accessing embeddings, REIS introduces a coarse-grained access scheme.
Specifically, REIS stores an address information entry for each region of the database in the internal DRAM. Each entry includes an integer index as the distinct signature of a database and the addresses of the first and last entries of the embedding and document regions. The coarse-grained access scheme enables database management in two ways. First, during database deployment, the storage system reserves two non-overlapped and consecutive regions and creates the address entries based on the size of a database before deploying the database to the storage system. In this way, we ensure the isolation of the database from other user data or databases. Second, during a database search operation, the storage system finds the starting embedding address of a database through the address entry to start the retrieval process. For each upcoming page read, the SSD controller infers the next address to read by incrementing the current address, instead of frequently invoking the address translation using the L2P mapping table. To ensure data integrity, REIS retains page-level FTL metadata, which contain essential information for operations such as refresh and wear-leveling. This metadata is used for: (i) writes during database initialization and (ii) periodic maintenance operations such as data refresh, which however are rare (e.g., once a year~\cite{micron9400}). After these operations, FTL metadata is flushed from the SSD's DRAM.

Coarse-grained access eliminates the need to maintain the page-level FTL for both regions of the database after deployment, conserving the valuable space of the internal DRAM for other operations (see Sec.~\ref{reis:isse}). For example, for a 1TB vector database that originally demands 1GB for page-level FTL \cite{zhou2021remap, han2019wal, wang2024learnedftl}, the maintenance cost for addressing is reduced to 21 bytes. Since REIS is designed with the aim of serving potentially many different RAG databases, we store the necessary information (i.e., the integer index of the database, the entries of the first/last entries in the embedding and document regions) in a small array in the SSD Controller's DRAM. This structure is called R-DB (\circled{A} in Fig.~\ref{fig:ov}) and serves as a record of deployed databases. A potential downside of coarse-grained access is that it requires the existence of a large contiguous block of storage, which may necessitate defragmentation operations during database deployment. However, this is an initial upfront overhead that can be amortized over time.

\subsection{ISP-Friendly ANNS Algorithms} 
\label{reis:impl-isp-friendly-anns}

\begin{figure}[htbp]
    \centering
    \includegraphics[width=\linewidth]{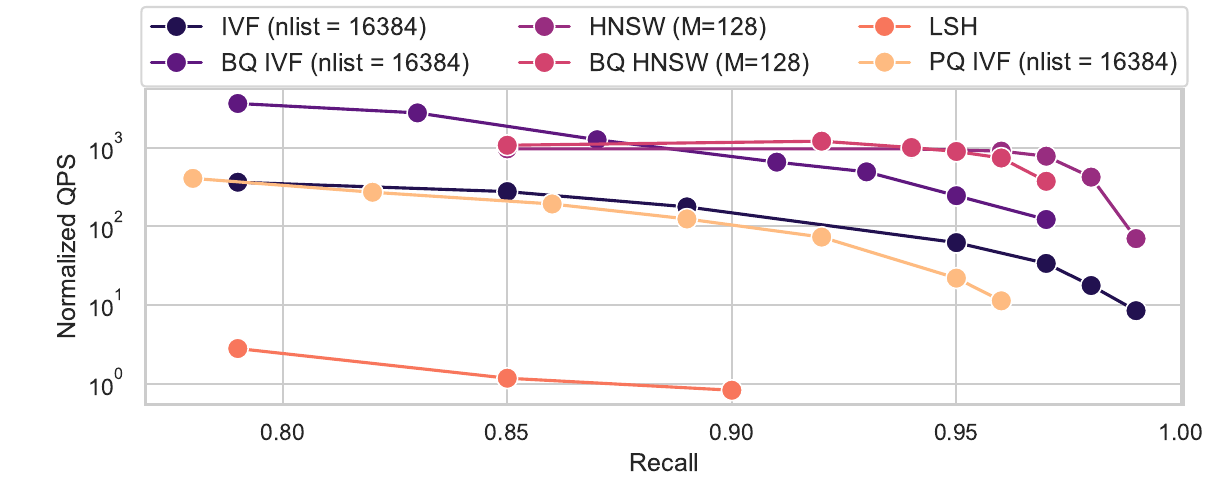}
    
    \caption{Comparison of ANNS algorithms in terms of throughput and recall running on CPU. For IVF, \textit{nlist} denotes the number of clusters for a dataset. For HNSW, \textit{M} denotes the number of neighbors for each vertex. }
    \label{fig:algorithm_selection}
\end{figure}

Apart from graph-based ANNS algorithms used by prior works~\cite{jayaram2019diskann,wang2023storage,malkov2018hnsw}, two other types of mainstream ANNS algorithms exist: cluster-based (e.g., Inverted File (IVF)~\cite{coster2002inverted,zobel2006inverted}) and hash-based algorithms (e.g., Locality Sensitive Hashing (LSH)~\cite{lshDasgupta}). With the aim of selecting the most suitable algorithms for our system, we perform a qualitative comparison, measuring throughput and recall on a CPU-based system (described in Table~\ref{tab:config}). Specifically, we compare the performance of IVF, HNSW, and LSH on the \textit{wiki\_en} dataset~\cite{dataset2} using the Cohere~\cite{reimers2022cohere} embedding model and the FAISS~\cite{douze2024faiss} library. We measure throughput in Queries per Second (QPS) and normalize it to that of exhaustive search. We first evaluate the performance of different implementations without quantization. Figure~\ref{fig:algorithm_selection} demonstrates that: (i) HNSW is the best performing base (i.e., without quantization) algorithm, (ii) both HNSW and IVF provide up to 0.99 recall, and (iii) LSH is the worst performing algorithm, with lower performance than exhaustive search (result) for recall values above $0.8$ ($1.2\times$ slower for Recall@10=0.9). 

Since ISP hardware has limited capabilities (e.g., lack of floating-point support~\cite{armr8}), ISP-based ANNS requires quantization. For this reason, in Fig.~\ref{fig:algorithm_selection} we also analyze the performance of IVF and HNSW when using Binary Quantization (BQ) and Product Quantization (PQ), combined with reranking~\cite{reimers2022cohere}. We make four key observations: (i) IVF recall remains high even with BQ (PQ) at 0.97 (0.96), (ii) PQ performs worse than BQ and even floating-point IVF, (iii) IVF throughput increases significantly with BQ, and (iv) HNSW throughput remains constant with BQ, while still outperforming IVF by approximately $3\times$. While these observations suggest that both HNSW and IVF are compelling options for ANNS-based RAG, graph-based algorithms (e.g. HNSW) feature irregular access patterns~\cite{faldu2019closer,faldu2020domain} that underutilize the internal bandwidth of the SSD, making them unsuitable for ISP. In contrast, IVF performs searches in contiguous data, exhibiting streaming access patterns. We thus select IVF as our algorithm of choice, and perform modifications to our database layout that support its execution.

\subsubsection{IVF-tailored Database Layout}
\label{reis:ivf}

In order to accelerate retrieval, REIS employs ANNS via the Inverted File (IVF) algorithm. As will be shown in Sec.~\ref{reis:isse}, REIS uses IVF with quantization and reranking, which requires storing data in both binary and INT8 precision.
To efficiently support IVF with these optimizations, we modify the database layout of Sec.~\ref{reis:dd} in three ways. First, we divide the embedding region into three sub-regions, one for storing cluster centroids, and two other regions for storing embeddings in binary and INT8 precision, respectively. 
Second, to facilitate IVF search operations, we create an array which serves as a record of all clusters. Each element of the array corresponds to an IVF cluster and contains: (i) the address of the cluster centroid, (ii) the index of the first and the last embedding within the cluster and (iii) a 8-bit tag associated with the cluster. We name this array R-IVF (\circled{B}) and store it in the SSD's DRAM, resulting in a memory footprint of $Number\_of_\_entries\times15B$.
Third, we extend the \textit{Embedding-Document Linkage} of Sec.~\ref{sub:embedding_doc_link} in two distinct ways. (I) In order to link binary embeddings to their INT8 counterparts for reranking, apart from the document address corresponding to each embedding, we also store the address of the INT8 embedding (RADR) in the OOB region. (II) For reasons that will become apparent in Sec.~\ref{reis:isse}, we store the 8-bit tag of the cluster in the OOB area of the page that contains the cluster centroid. 

Supporting IVF also requires allocating data structure in the SSD Controller's DRAM. Specifically, during IVF operations REIS maintains lists containing (i) clusters and (ii) embedding vectors as well as their distances from the query embedding. These structures are called \textit{Temporal Top Lists} (TTL) (\circled{C} in Fig.~\ref{fig:ov}) and as will be shown in Sec.~\ref{reis:isse} are employed in our In-Storage ANNS Engine.

\subsection{In-Storage ANNS Engine}
\label{reis:isse}

Prior ISP-based ANNS accelerators \cite{mailthody2019deepstore, liang2022vstore, wang2023storage} commonly integrate Multiple-Accumulate (MAC) units to compute Euclidean distance~\cite{anastasiu2015l2knng,arya1998optimal} for ANNS. 
Introducing such changes to the storage system creates (i) significant power and area overheads, and (ii) adoption issues due to the intrusive nature of such modifications. As explained in Sec. \ref{bg:anns}, there exist opportunities to reduce computational overhead of ANNS, while retaining accuracy. Recent studies \cite{muennighoff2022mteb, reimers2022cohere, shakir2024quantization, qdrant_binary,ms_binary} have shown that Binary Quantization (BQ) can achieve a recall of 96\%, due to the large dimensionality of text embeddings\embeddingDimensions{}. With REIS, our goal is to avoid the power and area overheads of prior designs. To this end, we design an In-Storage ANNS engine based on BQ which (i) utilizes only existing components within the SSD system to perform retrieval, (ii) exploits the plane-level, die-level, and channel-level parallelism of the storage system, and (iii) incorporates two major optimizations, distance filtering and pipelining.

\subsubsection{Search Process}

The search process for the Inverted File algorithm (IVF)~\cite{zobel2006inverted,coster2002inverted} consists of two steps, a coarse- and fine-grained search. First, in the coarse-grained search, REIS searches through all cluster centroids to identify those closest to the query embedding. To achieve this, REIS starts by reading and calculating the distance for all embeddings stored in one page. For each embedding, it then creates an entry consisting of the distance value (DIST), embedding (EMB), embedding address (EADR), and the associated tag (TAG). It sends these entries to a table, the Temporal Top List for Centroids (TTL-C), which resides in the SSD's DRAM. After filling the TTL-C for each page read, the embedded cores of the SSD controller execute a quickselect kernel~\cite{mahmoud1995analysis} on the distance numbers, identifying the entries that correspond to the \textit{N} nearest clusters to the query. Quickselect has an average time complexity of $O(N)$ and finds the \textit{k}-th smallest element in an unordered array, simultaneously selecting the $k$ smallest elements in the process without sorting them. At the same time, the storage system reads the next page of centroids and conducts distance computations to hide the latency of selection. Each iteration consists of (i) a page read, (ii) distance computations, and (iii) embedding selection, updating the TTL-C with the new closest clusters. After the last iteration, REIS selects the nearest clusters according to the finalized TTL-C. In the second step, REIS conducts a fine-grained search inside the clusters identified in the first step. The fine-grained search has two major differences compared to the coarse-grained search. (i) Instead of forming the TTL entry using TAG, for the fine-grained search, each TTL entry consists of DIST, EMB, RADR, and the address of the associated document (DADR). We name the table for the fine-grained search Temporal Top List for Embeddings (TTL-E). (ii) After the last iteration of selecting the \textit{k} nearest embeddings to the query, the storage system performs quicksort~\cite{hoare1962quicksort} to obtain a distance-ordered top-$k$ list for the query. 

\subsubsection{Retrieval architecture and execution}
\label{reis:retrieval}

\begin{figure}[h]
    \centering
    \includegraphics[width=\linewidth]{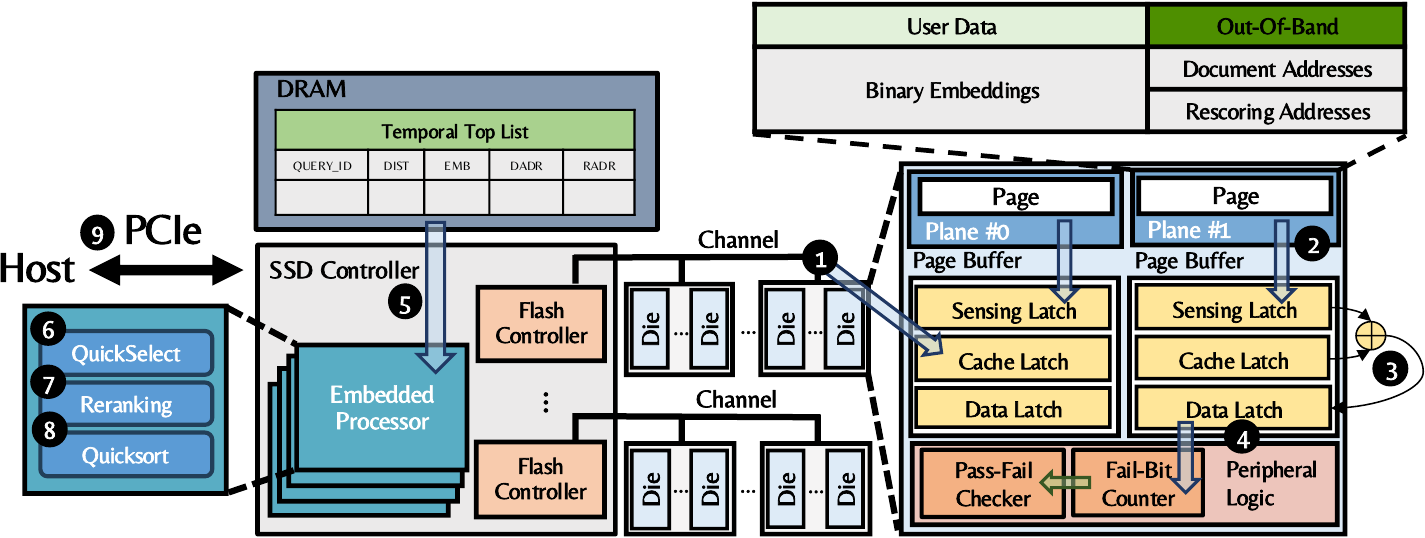}
    
    \caption{REIS's In-Storage ANNS Engine}
    \label{fig:reis}
\end{figure}

Document retrieval is performed by the \textit{ANNS engine}, which (i) receives the query embedding from the host system, (ii) computes the distance between the query embedding and database embeddings, and returns the top-\textit{k} results. Fig. \ref{fig:reis} breaks down REIS's execution flow in nine steps.

The execution flow begins with the reception of a new query by the storage system, which is placed in the SSD's DRAM and which triggers the execution of the ANNS kernel (steps \circled{2}-\circled{8}). The storage system first transfers the query embedding from the DRAM to the data buffer in each NAND Flash plane \circled{1} and then writes multiple copies of the data, filling the whole Cache Latch (CL). These copies are aligned to the database embeddings in order to enable bitwise operations, as will be described in step \circled{3}. We refer to this step as \textit{Input Broadcasting} (IBC). After IBC each CL holds \textit{N} duplicates, where \textit{N = Page\_Size /  Embedding\_Size}. In step \circled{2}, the storage system issues a page read command to each plane, loading a page of database embeddings to the Sensing Latch (SL). By performing an XOR operation between the CL (which stores the query embedding) and the SL (which now stores the database embeddings), and storing the result in the Data Latch (DL) \circled{3}, each plane calculates the bitwise difference of the query and the database embeddings. Next, in step \circled{4}, we employ the fail-bit counter~\cite{cho2024aero, micheloni2010inside, choi2020modeling} within the peripheral logic to measure the number of logical ones in the DL, which corresponds to the distance between the query and the database embeddings.

The data that is transferred out from the flash dies to the SSD Controller's DRAM changes depending on whether the steps \circled{1}-\circled{4} are executed during coarse- or the fine-grained search. For coarse-grained search, the ANNS engine transfers the (i) the embedding vector (EMB), (ii) its calculated distance (DIST), and (iii) the tag of the cluster that this embedding belongs to, forming a single entry. For fine-grained search, instead of transferring the tag of the cluster, the ANNS engine transfers (iv) the addresses of the INT8 version of the embeddings (RADR), and (v) the correlated document chunk address (DADR). Steps \circled{2}-\circled{4} are repeated until the whole database is searched. The SSD controller retrieves distance numbers from the TTL \circled{5} and performs quickselect~\cite{hoare1961qselect} using the embedded core \circled{6}, selecting the $10k$ embeddings closest to the query. In step \circled{7}, the embedded core of the SSD controller executes the reranking kernel~\cite{yamada2021efficient, reimers2022cohere, qdrant_binary}. Reranking performs a \textit{costlier} but more accurate search on the subset of data elements that are selected by ANNS. Rerankers usually (i) employ cross-encoder models that accurately calculate the similarity between queries and document chunks~\cite{nogueira2019multi, chen2024bge}, or (ii) recalculate distances with higher precision (e.g., INT8)~\cite{shakir2024quantization}. REIS uses the second approach: ANNS is performed using Binary Quantization, while reranking is performed using INT8 embeddings. For reranking, the embedded core first fetches the top-$10k$ embeddings from the INT8 embedding region using the RADR. It then recalculates the distances in INT8 precision and sorts them using quicksort~\cite{hoare1962quicksort} \circled{8} to finally select the top-$k$ embeddings, which ends the search process. Once the ANNS search is completed, the ANNS engine executes \textit{document identification} to find relevant document chunks according to the DADR of the top-\textit{k} results  and transfers them to the host system \circled{9} for generation.

\noindent\textbf{Exploiting SSD Parallelism.}
As described in Sec.~\ref{reis:retrieval}, REIS uses the buffers and the peripheral logic within the planes and the dies of the storage system in order to perform distance computations. This approach allows multiple simultaneous XOR and bit-counting operations across planes and dies, exploiting the available parallelism within the storage system. Once these computations have been performed, the flash channels of the storage system collectively provide massive internal bandwidth (e.g., 9.6 GB/s bandwidth for an 8-channel system with 1.2 GB/s bandwidth per channel~\cite{cho202130}), which can efficiently transfer entries from the flash dies to the SSD controller's DRAM by leveraging the channel-level parallelism.

\noindent\textbf{Fine-grained Embedding Access.}
To ensure fine-grained access to each  embedding, REIS introduces \emph{Mini-Pages} for addressing. REIS composes a \emph{Mini-Page} address by appending an offset to the original physical page address, filling each page with as many embeddings as possible (e.g., 128 binary 1024-dimension embeddings per 16KB page, leading to a 7-bit offset for the \emph{Mini-Page} address). During execution of the ANNS engine, REIS performs retrieval using the \emph{Mini-Page} address as the embedding address (EADR) for each entry in the TTL.

\subsubsection{Distance Filtering.}
\label{reis:distance-filtering}
We experimentally find that, for each query, a significant fraction of document chunks within the database are irrelevant (i.e., the distance between their embeddings and the query embedding is very large). For example, various retrieval tasks, such as fact-checking~\cite{thorne2018fever}, retrieve only $1.2$-$3.0$ relevant document chunks per query on average from the BEIR datasets~\cite{thakur2021beir}. To avoid forwarding irrelevant data to the SSD controller, we employ distance filtering, which discards database embeddings when their distance from the query embedding exceeds a certain threshold. By discarding highly irrelevant queries, distance filtering (i) conserves SSD channel bandwidth, and (ii) reduces the number of entries that the SSD controller has to select and sort.

We introduce a modification to step \circled{4} with which we apply distance filtering to the ANNS kernel. To determine suitable thresholds, we perform filtering experiments on 4 BEIR~\cite{dataset_beir} datasets targeting different retrieval tasks: \textit{HotpotQA}~\cite{yang2018hotpotqa}, \textit{NQ}~\cite{kwiatkowski2019natural}, \textit{FEVER}~\cite{thorne2018fever}, and \textit{Quora}~\cite{dataset_quora}. We make two observations: First, for \textit{HotpotQA} we can filter out $99\%$ of the documents and still retrieve the \textit{k}=10 most relevant ones for each query. Second, the choice of filtering threshold only weakly depends on the dataset size. For $k=10$, the threshold would only be $1.6\%$ higher for the biggest dataset, \textit{FEVER} compared to the smallest, \textit{Quora}. We conclude that (i) distance filtering significantly reduces the number of candidate embeddings and thus computation, and (ii) it is possible to employ one filtering threshold for effectively filtering datasets with different sizes.

We implement distance filtering using the comparator logic within the flash dies (i.e., the pass/fail checker)~\cite{cho2024aero, micheloni2010inside, choi2020modeling}, which compares distance numbers with a pre-defined threshold. Each embedding whose distance (DIST) value is below the threshold is transferred to the SSD's DRAM for further processing.

\subsubsection{Pipelining.}

To further accelerate RAG retrieval, REIS exploits three pipelining opportunities within the storage system. First, REIS leverages the widely implemented {\em Read Page Cache Sequential} mode~\cite{micheloni2010inside}, inside the flash chips, to overlap operations between two iterations of steps \circled{2}-\circled{4}. 
Specifically, during step \circled{4}, after the PL transfers its data to the DL for readout, it can immediately read the next page. Second, REIS overlaps distance calculation on the NAND Flash dies with kernel execution on the embedded cores. According to our evaluation, a single core can efficiently run Quicksort and reranking without stalling the pipeline. Therefore, REIS only uses one core for Quicksort and reranking, while the other cores (e.g., 3 out of 4~\cite{samsung990pro, samsung980pro}) are still available for regular SSD operations.
Third, during IBC (see Sec.~\ref{reis:retrieval}), REIS enables all planes per die to receive the input query from the die I/O simultaneously, an optimization that we name Multi-Plane IBC (MPIBC). 
This reduces the IBC latency by a factor equivalent to the number of planes per die. We assume the plane selection is handled by a dedicated Multiplexer logic within the die periphery. Therefore, enabling MPIBC requires raising the select signal for all planes together so that they can receive the input query embedding concurrently.

\subsection{System Integration of REIS}
\label{reis:sys}

To enable communication with the host, REIS introduces an Application Programming Interface (API) that defines RAG-specific extensions to the NVM command set~\cite{nvmcommands}. Similarly, to support the operations described in Sec.~\ref{reis:isse}, REIS extends the NAND flash command set with commands that enable communication between the controller and the flash dies.

\subsubsection{Application Programming Interface}
\label{reis:api-design}
REIS specifies a high-level API for the host system to perform the indexing and the retrieval stage of the RAG workflow. To achieve this, we extend the NVM command set~\cite{nvmcommands} with custom REIS operations. The specification provides a range ($80h$-$FFh$) in the opcode values for vendor-specific commands, which are adequate for implementing all REIS operations. 
To perform indexing, the host system issues \textit{DB\_Deploy()} (or \textit{IVF\_Deploy()}) to the SSD. REIS reserves the required space in the NAND Flash memory according to the API and performs de-fragmentation operations to create a physical contiguity. 
It then waits for the host to write the database content to the DRAM, which it subsequently writes to storage as explained in Sec. \ref{reis:dd}.
When REIS receives \textit{Search()} (or \textit{IVF\_Search()}) from the host system, it performs retrieval and returns a \textit{done} signal once it has identified the document chunks to be retrieved. Once the host system acknowledges the signal, the storage system starts to transfer the identified document chunks to the host system.
Table~\ref{tab:api} describes each API command.

\begin{table}[h!]

\caption{REIS Application Programming Interfaces}
\small
\label{tab:api}
    \centering
    \begin{tabularx}{\linewidth}{l|X}
        \toprule
        \textbf{API Commands} & \textbf{Description} \\
        \midrule\midrule
        \textit{DB\_Deploy(DB, D\textsubscript{id}, N)} &
            \begin{tabular}[c]{@{}X@{}}Write the \textit{N}-entry database \textit{DB}, with ID \textit{D\textsubscript{id}} to storage.\end{tabular}\\
        \midrule
        \textit{IVF\_Deploy(DB, D\textsubscript{id}, N, CI)} &
            \begin{tabular}[c]{@{}X@{}}Write the \textit{N}-entry  IVF-based database \textit{DB}, with ID \textit{D\textsubscript{id}} to storage. \textit{CI} contains information on the IVF clusters.\end{tabular}\\
        \midrule
        \textit{Search(Q, Q\textsubscript{id}, D\textsubscript{id}, k)} &
            \begin{tabular}[c]{@{}X@{}}Perform a top-\textit{k} search for a batch of queries \textit{Q}, indexed by \textit{Q\textsubscript{id}}, in the database with ID \textit{D\textsubscript{id}}. \end{tabular} \\
        \midrule
        \textit{IVF\_Search(Q, Q\textsubscript{id}, \textit{D\textsubscript{id}}, k, R)} &
            \begin{tabular}[c]{@{}X@{}}Perform a top-\textit{k} IVF search for a batch of queries \textit{Q}, indexed by \textit{Q\textsubscript{id}}, in the database with ID \textit{D\textsubscript{id}}. The target accuracy is \textit{R}.\end{tabular} \\
        \bottomrule
    \end{tabularx}
\end{table}

\subsubsection{NAND Flash Command Set}
\label{reis:flash-commands}

REIS adds new commands to the NAND flash die control logic to support the operations of the in-storage ANNS engine for retrieval tasks. To enable this, the controller first receives the previously described API commands and translates them into the flash command set. It then issues the flash commands to the flash dies to perform the necessary operations. The control logic within each flash die is a finite-state machine, which receives the commands and uses them to control the peripheral logic in the flash array. Table~\ref{tab:isa} describes the NAND flash command set extensions for querying the database.

\begin{table}[h!]

\caption{NAND Flash Command Set Extensions}
\label{tab:isa}
    \centering
    \small
    \begin{tabularx}{\linewidth}{l|X}
        \toprule
        \textbf{ISA Format} & \textbf{Description} \\
        \midrule\midrule
        \texttt{IBC Q\_EMB} &
            \begin{tabular}[c]{@{}X@{}}Send a copy of the query (\texttt{Q\_EMB}) to each page buffer of the NAND Flash memory. (\textit{Input Broadcasting})\end{tabular} \\
        \midrule
        \texttt{XOR ADR\_P} &
            \begin{tabular}[c]{@{}X@{}}Perform the XOR operation between PL and CL of a plane (addressed by \texttt{ADR\_P}).\end{tabular} \\
        \midrule
        \texttt{GEN\_DIST EADR} &
            \begin{tabular}[c]{@{}X@{}}Compute the distance for a database embedding stored at address \texttt{EADR}.\end{tabular} \\
        \midrule
        \texttt{RD\_TTL EADR} &
            \begin{tabular}[c]{@{}X@{}}Transfer the TTL entry for the embedding stored at \texttt{EADR} to the SSD DRAM.\end{tabular} \\
        \bottomrule
    \end{tabularx}
\end{table}

\section{Methodology}

\noindent\textbf{Evaluated System Configurations.}
We evaluate REIS on two SSD configurations, \textbf{REIS-SSD1} and \textbf{REIS-SSD2}, based on two commercial SSD products, Samsung PM9A3 \cite{pm9a3} and Micron 9400 \cite{micron9400}. These SSDs focus on low cost and high performance, respectively.
As a baseline for document retrieval, we use a high-end server equipped with an AMD EPYC 9554 CPU~\cite{amdepyc} and a Samsung PM9A3 SSD~\cite{pm9a3}. Table~\ref{tab:config} provides the properties of our SSDs and the baseline CPU system (\textbf{CPU-Real}). To highlight the improvements stemming from our database layout and In-Storage Processing, we first compare REIS and CPU-Real using brute force search (BF). We then compare REIS and CPU-Real on Approximate Nearest Neighbor Search. Since (i) the loading time makes up the biggest fraction of the execution time (see Sec.~\ref{mot:limitations}), and (ii) HNSW indexes take up significantly more space than IVF ones, IVF outperforms HNSW when loading time is taken into account. We evaluate both REIS and CPU-Real with the IVF algorithm using BQ and reranking, provided by the FAISS library~\cite{douze2024faiss}, sweeping the accuracy of IVF from 0.98 down to 0.9 $Recall@10$. In order to perform a sensitivity study, we introduce \textbf{No-OPT} as a baseline, a REIS configuration that uses the In-Storage ANNS Engine without DF, PL, and MPIBC. To quantify the performance overheads stemming from ANNS only, we introduce an additional comparison point based on the CPU baseline, which incurs zero overheads from data movement due to storage I/O, called \textbf{No-I/O}. We additionally compare REIS to two state-of-the-art designs, NDSearch~\cite{wang2023storage} and ICE~\cite{hu2022ice}, which use graph-based and cluster-based ANNS, respectively. To ensure a fair comparison we make the appropriate modifications to our experimental methodology whenever required.

\noindent\textbf{Performance \& Energy Evaluation.}
Our SSD operation model and parameters are based on Flash-Cosmos~\cite{park2022flash} while the internal SSD DRAM is modeled using CACTI7~\cite{balasubramonian2017cacti}. We use Zsim~\cite{sanchez2013zsim} and Ramulator~\cite{kim2015ramulator,ramulator-github} to simulate the embedded SSD controller cores. We model SSD power consumption based on a commodity product~\cite{samsung980pro} and real chip characterization results from Flash-Cosmos~\cite{park2022flash}. The power of the SSD's internal DRAM and that of the embedded cores are also derived from CACTI7~\cite{balasubramonian2017cacti} and the characteristics of a commodity embedded SSD controller processor~\cite{armr8}, respectively.
We measure the power of CPU-Real using AMD $\mu$Prof \cite{amduprof} for the CPU and a DDR4 model~\cite{micronddr4, ghose2019demystifying} for DRAM.

\noindent\textbf{Evaluated Datasets.}
We evaluate two datasets from an information retrieval benchmark~\cite{thakur2021beir}, \textit{NQ} and \textit{HotpotQA}, a public dataset based on wikipedia~\cite{dataset2} (\textit{wiki\_full}) and its English subset (\textit{wiki\_en}). For the comparison to NDSearch~\cite{wang2023storage} we use two billion-scale datasets that were used to evaluate NDSearch, \textit{SIFT1B} and \textit{DEEP1B}~\cite{simhadri2022results}.  

\begin{table}[h!]
\caption{Evaluated System Configurations}
\label{tab:config}
    \centering
    \small
    \begin{tabular}{l|l}
        \toprule
        \textbf{System} & \textbf{Configuration}\\
        \midrule
        \midrule
        \textbf{CPU-Real}&
          \begin{tabular}[c]{@{}l@{}}CPU: 2 sockets, 128 cores, 3.1GHz \cite{amdepyc};\\ DRAM: 1.5TB DDR4 \cite{micronddr4}; SSD: PM9A3 \cite{pm9a3}\end{tabular} \\ 
        \midrule
        \textbf{REIS-SSD1} &
          \begin{tabular}[c]{@{}l@{}}8 channels; 16 512Gb dies/channel; 2 planes; \\
          1.2 GB/s channel bandwidth; \\
          22.5µs tR (ESP-SLC) \cite{park2022flash}; \\
          Embedded Cores: Cortex R8 \cite{armr8}; 4 cores;\end{tabular} \\ 
        \midrule
        \textbf{REIS-SSD2} &
          \begin{tabular}[c]{@{}l@{}}16 channels; 8 512Gb dies/channel; 4 planes; \\
          2.0 GB/s channel bandwidth; \\
          22.5µs tR (ESP-SLC) \cite{park2022flash}; \\ 
          Embedded Cores: Cortex R8 \cite{armr8}; 4 cores;\end{tabular} \\ 
          \bottomrule
    \end{tabular}
\end{table}

\section{Evaluation}

We evaluate the effectiveness of REIS compared to different baselines.
First, we evaluate the effectiveness of REIS at improving the performance and energy efficiency of the retrieval stage of the RAG pipeline. Second, we evaluate the effect of REIS on the performance of the end-to-end RAG pipeline. Third, we conduct a sensitivity study to analyze the effect of different optimization techniques in REIS. Fourth, we compare REIS to two prior works~\cite{hu2022ice,wang2023storage} that use cluster- and  graph-based ANNS algorithms, respectively.

\subsection{Retrieval Performance \& Energy Efficiency}
\label{eval:isp}

\begin{figure*}[ht]
    \centering
    \includegraphics[width=\textwidth]{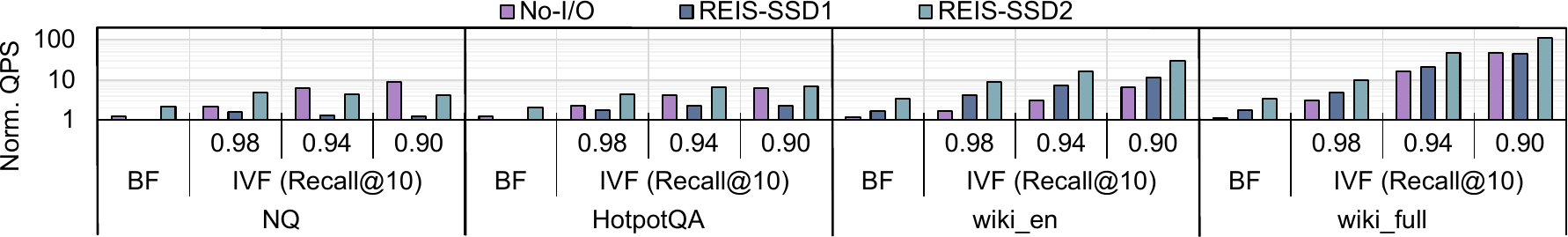}
    
    \caption{Performance (QPS) normalized to CPU-Real}
    \label{fig:perf}
\end{figure*}

\noindent\textbf{Performance.}
Figure~\ref{fig:perf} shows the performance of REIS, measured in Queries-per-Second (QPS) and normalized to CPU-Real. We make three observations. First, REIS-SSD1 and REIS-SSD2 improve performance over CPU-Real by an average of $13\times$ with a maximum of $112\times$, demonstrating the benefit of alleviating the I/O bottleneck of the RAG retrieval process. Second, REIS-SSD1 and REIS-SSD2 outperform No-I/O by an average of $1.8\times$ with a maximum of $5.3\times$ due to the massive internal parallelism of storage systems that REIS exploits. Third, REIS-SSD2 provides a $2.6\times$ average speedup over REIS-SSD1, with a maximum of $3.2\times$, reflecting the benefits of higher channel counts ($2\times$) and channel bandwidth ($1.7\times$).

\begin{figure*}[ht]
    \centering
    \includegraphics[width=\textwidth]{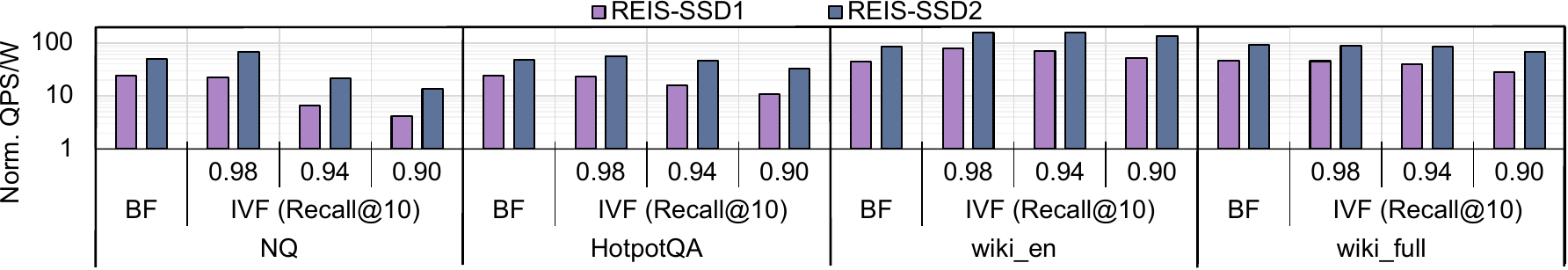}
    
    \caption{Energy efficiency (QPS/W) normalized to CPU-Real}
    \label{fig:energy}
\end{figure*}

\noindent\textbf{Energy Efficiency.}
Figure~\ref{fig:energy} presents the energy efficiency (QPS/W) of REIS normalized to CPU-Real. We make two observations. First, REIS-SSD1 and REIS-SSD2 improve energy efficiency over CPU-Real by $55\times$ on average and up to $157\times$. This improvement in energy efficiency fundamentally stems from the $29.7\times$ lower power consumption of SSDs compared to the CPU baseline on average. Second, REIS-SSD2 provides $2.2\times$ higher energy efficiency over REIS-SSD1 on average, with a maximum of $2.6\times$. This improvement in energy efficiency is similar to REIS-SSD2's performance improvement over REIS-SSD1, suggesting that most of the energy efficiency gains stem from the higher throughput of SSD2's design. 

\subsection{End-to-End Performance Analysis}
\label{eval:rag-pipeline}

Table~\ref{tab:latency_breakdown_comparison} breaks down the latency of different stages of the RAG pipeline on REIS-SSD1 and on a CPU-based system using binary quantization (i.e., the same system as in Fig.~\ref{fig:mot2}). Similarly to our analysis in Fig.~\ref{fig:mot2}, we use the \textit{HotpotQA} and \textit{wiki\_en} datasets. Since REIS performs retrieval within the storage system, it does not perform the \textit{Dataset Loading} step that transfers data to the host's DRAM.  We observe that REIS reduces the combined latency of \textit{Dataset Loading} and \textit{Search} from 20.3\%-69.3\% down to 0.02\%-0.15\%, which demonstrates that REIS efficiently eliminates the data movement bottleneck of RAG retrieval. When using REIS, \textit{Generation} accounts for 92\% of the total time, which demonstrates that LLM inference is now the new bottleneck. Overall, REIS reduces the average end-to-end latency by $1.25\times$ and $3.24\times$ on \textit{HotpotQA} and \textit{NQ}, respectively.

\setlength{\extrarowheight}{1.4pt} 
\begin{table}[b]
\caption{RAG Latency Breakdown for REIS and the CPU-based system with Binary Quantization of Fig.~\ref{fig:mot2}.}
\label{tab:latency_breakdown_comparison}
\footnotesize
\centering
\begin{tabular}{l|c|c||c|c|}

\Xcline{2-5}{\arrayrulewidth}
& \multicolumn{2}{c||}{\textbf{HotpotQA}} & \multicolumn{2}{c|}{\textbf{NQ}} \\ 

\Xhline{2\arrayrulewidth}
\rowcolor[HTML]{EFEFEF}
\multicolumn{1}{|l|}{\textbf{Latency contribution (\%)}} & \textbf{REIS}&\textbf{CPU+BQ}&\textbf{REIS}&\textbf{CPU+BQ}\\ \hline \hline
\multicolumn{1}{|l|}{Embedding Model Loading} & 3.26 & 2.61 & 3.26 & 1.01 \\ \hline

\multicolumn{1}{|l|}{Encoding} & 0.58 & 0.46 & 0.58 & 0.18 \\ \hline

\multicolumn{1}{|l|}{Dataset Loading} & N/A & 20.0 & N/A & 67.3 \\ \hline

\multicolumn{1}{|l|}{Search (and retrieval for REIS)} & 0.02 & 0.29 & 0.15 & 2.00 \\ \hline

\multicolumn{1}{|l|}{Generation Model Loading} & 4.16 & 3.32 & 4.16 & 1.28\\ \hline

\multicolumn{1}{|l|}{Generation} & 92.0 & 73.0 & 92.0 & 28.0\\ \hline \hline

\rowcolor[HTML]{EFEFEF}
\multicolumn{1}{|l|}{\textbf{End-to-End Latency (s)}} & 18.97 & 23.79 & 19.0 & 61.69 \\ 

\Xhline{\arrayrulewidth}
\end{tabular}
\end{table}

\subsection{Sensitivity Study}
\label{eval:abl}

\begin{figure}[htbp]
    \centering
    \begin{subfigure}{0.43\linewidth}
        \includegraphics[width=\linewidth]{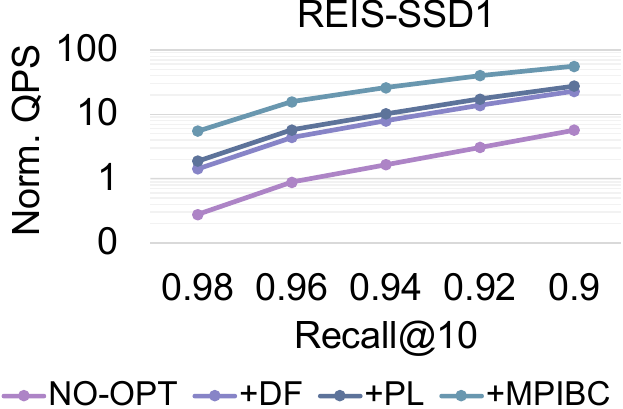}
    \end{subfigure}
    \hspace{0.05\textwidth}
    \begin{subfigure}{0.43\linewidth}
        \includegraphics[width=\linewidth]{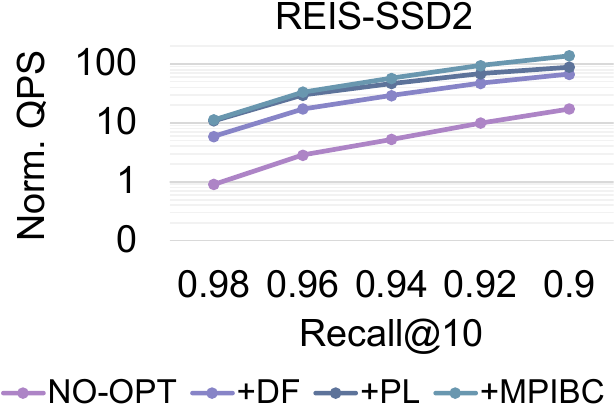}
    \end{subfigure}
    \caption{Effects of different REIS optimizations on throughput (normalized to CPU-Real), evaluated on dataset~\cite{dataset2}.}
\label{fig:abl}
\end{figure}

Fig.~\ref{fig:abl} presents a sensitivity study of all proposed optimizations introduced by REIS, i.e., Distance Filtering (DF), Pipelining (PL) and Multi-Plane Input Broadcasting (MPIBC) on top of No-OPT. We choose \textit{wiki\_full}~\cite{dataset2} as the dataset to analyze and normalize results (i.e., QPS) to the performance of the CPU-Real. We make three observations. First, among all proposed optimizations, DF contributes the most to the speedup over No-OPT by an average of 4.7$\times$ and 5.7$\times$ and a maximum of 5.1$\times$ and 6.5$\times$ for REIS-SSD1 and REIS-SSD2, respectively. The main source of this speedup is that filtering out embeddings with large distances inside each NAND flash die significantly reduces (i) unnecessary data movement to the SSD controller's DRAM, and (ii) the amount of data input to the Quickselect kernel.
Second, the benefit from PL increases for SSDs with higher internal bandwidth due to more channels and higher I/O rate. Specifically, in SSDs with high internal bandwidth (e.g., the 32GB/s of bandwidth for REIS-SSD2), pipelining can completely overlap (i) reading a new page, and (ii) transferring out the filtered TTL entries from the NAND flash dies to the SSD's internal DRAM. 
Third, the benefit from MPIBC increases for SSDs with more planes per die. 
Specifically, the average speedup of DF+PL+MPIBC over DF+PL is 6\% and 26\% for REIS-SSD1 and REIS-SSD2. 

\subsubsection{Comparison with REIS-ASIC}
To quantify the performance loss due to not using ESP (thus requiring ECC which incurs data transfers to the SSD controller), we compare REIS against a new scheme, REIS-ASIC, which: (i) instead of ESP, uses ECC performed by the SSD controller, (ii) performs all other operations using an ideal ASIC with no computational overhead but (iii) requires that all data be transferred to the controller. REIS-ASIC experiences a slowdown between $4.1\times$-$5.0\times$  ($3.9\times$-$6.5\times$) for SSD-1 (SSD-2), across all recall values and datasets, due to the data movement overheads introduced by the data transfers due to not using ESP.

\subsection{Comparison to Prior Works}
\label{eval:algo}

We compare the performance of REIS to two state-of-the-art ISP-based ANNS accelerators, ICE~\cite{hu2022ice} and NDSearch~\cite{wang2023storage}, which use cluster- and graph-based algorithms, respectively.

\begin{figure*}[ht]
    \centering
    \includegraphics[width=\textwidth]{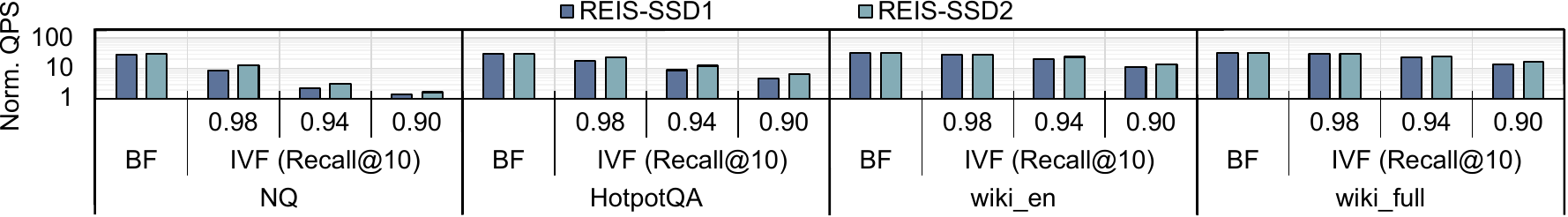}
    
    \caption{Speedup of REIS over  ICE~\cite{hu2022ice}.}
    \label{fig:ice-comp}
\end{figure*}
\noindent\textbf{Comparison to ICE.}
 Fig.~\ref{fig:ice-comp} shows the speedup of REIS compared to ICE~\cite{hu2022ice}, a state-of-the-art ISP scheme for vector similarity search. When using brute force (BF), REIS achieves a speedup greater than $10\times$ across all configurations. For IVF, the speedup increases with higher recall values, demonstrating superior performance to that of ICE. Specifically, across all datasets with SSD-2, REIS outperforms ICE by an average of $7.1\times$ ($22.9\times$) at 0.90 (0.98) recall@10. We also perform a comparison to ICE-ESP, an idealistic implementation of ICE that does \emph{not} require ECC, but still uses 4-bit quantization (not shown in Fig.~\ref{fig:ice-comp}). Even compared to ICE-ESP, REIS achieves a geomean speedup of $3.85\times$ ($3.92\times$) in BF for SSD-1 (SSD-2). When configured to target 0.9 recall@10 using IVF, REIS achieves $2.08\times$ ($2.29\times$) higher performance over ICE-ESP, a number that rises to $2.84\times$ ($3.18\times$) for 0.98 recall@10 for SSD-1 (SSD-2).

\begin{figure}[htbp]
    \centering
    \includegraphics[width=0.9\linewidth]{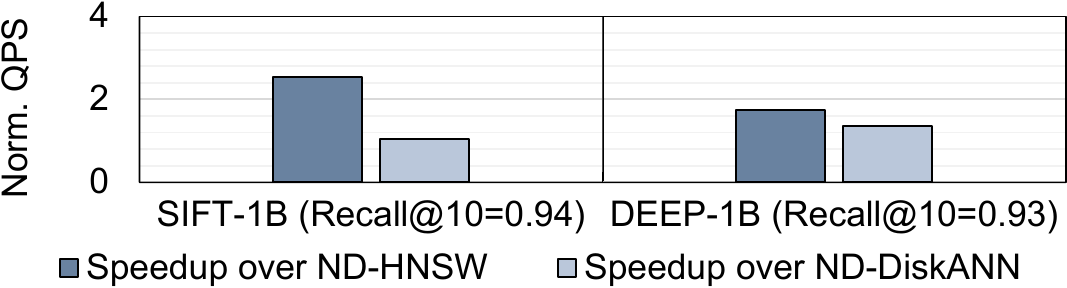}
    
    \caption{Performance comparison of REIS and NDSearch.}
    \label{fig:reisndsearch}
\end{figure}

\noindent\textbf{Comparison to NDSearch.} Fig.~\ref{fig:reisndsearch} compares the performance of REIS using IVF~\cite{douze2024faiss}, against NDSearch using HNSW~\cite{malkov2018hnsw} and DiskANN~\cite{jayaram2019diskann}.
We perform this comparison using two billion-scale datasets, \textit{SIFT-1B} and \textit{DEEP-1B}~\cite{simhadri2022results}, with  0.94 and 0.93 Recall@10, respectively.
We normalize the throughput of REIS to that of NDSearch with HNSW and DiskANN and observe that it outperforms NDSearch by an average of 1.7$\times$ with a maximum of 2.6$\times$.

\section{Discussion}
\label{reis:flex}

In this section, we discuss
potential extensions and optimizations to REIS. First, we discuss augmenting REIS with filtered search on user-defined metadata.  Second, we address the impact of REIS on typical SSD management operations and lifetime. Third, we provide alternative implementations for REIS's embedding-document linkage which alleviate the logical to physical contiguity requirements.

\subsection{Metadata Filtering}

To improve generation quality modern LLM serving frameworks~\cite{LlamaIndex,LangChain} incorporate \emph{metadata filtering}~\cite{poliakov2024multi,wang2023learning,wang2024blendfilter} to RAG retrieval. Metadata filtering augments database entries with information such as timestamps, author information, or other relevant metadata that can be used during the search process to improve document retrieval. REIS could potentially be enhanced with this feature by storing the metadata of each embedding in reserved NAND flash memory (i.e., in the OOB region~\cite{yang2024lbz}). 

To perform metadata filtering in a read-only database~\cite{poliakov2024multi, jin2024flashrag}, this enhanced version of REIS: (i) assigns a corresponding metadata tag (an integer number) to each embedding  and (ii) places the tag in the OOB area during database deployment. During RAG retrieval, REIS receives the query embedding alongside a metadata tag and compares it to the tags of each database embedding, using the existing
approach for calculating the embedding distance. Before performing the subsequent retrieval steps, REIS checks the result of the metadata computation, filtering out results that do not match. For continuously updated databases providing real-time knowledge retrieval~\cite{gade2024s, cheng2024unified, chen2021dataset, kasai2024realtime}, REIS (i) periodically creates new databases to store new information at a predefined frequency (e.g., every hour), (ii) treats each sub-database as a normal database tagged with an individual timestamp, (iii) maintains an entry for each database in the internal DRAM, including the database address and the timestamp. When the host sends a query with a requested time, REIS identifies the corresponding databases to be searched by first comparing the requested time with the timestamps stored in the internal DRAM and then performs search and retrieval operations within the identified databases.

\subsection{Implications on the Storage System}
\label{mech:ssd-impact}
\noindent\textbf{Typical SSD operations.} While REIS is primarily designed to accelerate RAG, it also serves as a conventional storage system. As such, the SSD controller must handle routine maintenance tasks, such as data refresh and garbage collection~\cite{yang2019reducing,yan2017tiny,kang2017reinforcement}. To ensure uninterrupted execution of maintenance operations, we (i) confine REIS to only one of the embedded cores of the SSD and (ii) prioritize maintenance tasks over RAG operations when all cores are needed for maintenance. Since REIS primarily targets read-intensive RAG workloads, write operations are expected to be infrequent, making full core utilization a rare occurrence. To simplify the design, REIS operates exclusively in either RAG-mode or normal SSD mode at any given time. To switch between the two modes, it is necessary to load the necessary FTL data (coarse-grained for RAG (see Sec.~\ref{reis:coarsega}), fine-grained for normal operations). Since REIS exclusively operates in one of the two modes, performance of normal read/write operations from the host remains unaffected.

\noindent\textbf{Impact on SSD Lifetime.} Although REIS disables ECC in the SLC partition to support in-die logic operations, this does not reduce SSD lifetime for two reasons. First, using SLC-mode instead of MLC inherently increases the distance between threshold voltages, enhancing flash memory cell reliability. Second, REIS employs ESP for the SLC partition, which achieves a 0 BER~\cite{park2022flash}, in a worst-case scenario, (i.e., 1-year retention time, 10k Program/Erase cycles)~\cite{park2022flash}.

\noindent\textbf{Contiguity Requirements.} Coarse grained access (i.e., the lightweight L2P mapping scheme of Sec.~\ref{reis:coarsega}) requires the existence of contiguous unallocated physical space. In order to further reduce (i) the memory footprint, and (ii) translation overheads stemming from L2P metadata, REIS also uses the same contiguity-based approach in the document region of the database. An alternative approach, which does not require contiguity in the document region, would be to link embeddings to the physical addresses of their corresponding document chunks via the OOB area, enabling document chunks to be placed anywhere in storage. However, this approach introduces additional complexity as it entails updating the physical address in the OOB region whenever the documents are remapped to another region of the SSD (e.g., during updates).

\section{Related Work}

To our knowledge, REIS is the first system based on In-Storage Processing (ISP) that accelerates the retrieval stage of Retrieval-Augmented Generation (RAG).  We have already qualitatively and quantitatively compared REIS to two existing state-of-the-art ISP-based ANNS accelerators~\cite{hu2022ice,wang2023storage} in Section~\ref{eval:algo}. In this section, we discuss works that improve RAG from other perspectives and relevant works for Nearest Neighbor Search Acceleration.

\subsection{RAG Enhancements}
Prior work has proposed various optimizations to the RAG pipeline.
RQ-RAG~\cite{chan2024rq}, a representative prompt engineering~\cite{gao2022precise, wang2023query2doc, kim2023tree, chan2024rq} method, decomposes complex queries and disambiguates queries with more than one possible interpretation.
Small-to-Big Retrieval \cite{yang2023small2big}, an improved document chunking strategy~\cite{raina2024question, theja2024eval, textsplitters}, uses small document chunks for the retrieval search and returns bigger chunks covering the same context. 
Hybrid approaches incorporate dense retrieval with sparse retrieval to capture both semantic and lexical similarity between query and documents~\cite{wang2023rap, lu2022reacc}, or combine database search with web search when the knowledge base cannot provide relevant information~\cite{yan2024corrective}. 

\subsection{Nearest Neighbor Search Acceleration} \label{ann_related}

Due to the widespread adoption of ANNS to billion-scale recommendation systems~\cite{zhang2018visual, huang2020embedding, zhang2022uni, li2021embedding, gan2023binary, fu2017fast}, recent works have proposed dedicated libraries~\cite{douze2024faiss, ANNScaling, sun2020scann} and optimized algorithms \cite{malkov2014approximate, malkov2018hnsw, munoz2019hierarchical, vecchiato2024learning, bruch2024optimistic} to improve its performance. 
These works improve the performance of ANNS through various optimizations for processor-centric systems. Since these optimizations target processor-centric systems, they cannot overcome the I/O data movement bottleneck that REIS aims to alleviate.

Various ANNS hardware accelerators~\cite{zeng2023df, liang2022vstore, wang2023storage, xu2023proxima, khan2024bang, jang2023cxl, groh2022ggnn, ootomo2023cagra, tian2024scalable, zhu2023processing, ren2020hm} leverage approaches such as memory expansion \cite{jang2023cxl, ren2020hm} and multi-node parallelism. \cite{zeng2023df, groh2022ggnn, tian2024scalable}. Processing-in-Memory techniques (PIM) have also been explored for accelerating Nearest Neighbor Search. For example, \cite{quinn2025accelerating} proposes a CXL-based device that places vector product accelerators near LPDDR memory, aiming to improve the performance of Exact Nearest Neighbor Search (ENNS). In~\cite{qin2024robust}, Qin et al. leverage the properties of Non-Volatile Memory technologies to perform matrix-vector multiplication in the analog domain and accelerate RAG pipelines in edge devices.
Despite performance improvements, DRAM-based approaches either fail to fundamentally address the I/O data movement bottleneck from storage or incur significant costs to serve large datasets.

\section{Conclusion}

We introduce REIS, a new retrieval system tailored to Retrieval-Augmented Generation based on In-Storage Processing. REIS improves performance and energy efficiency, by leveraging the existing computational resources within the storage system. REIS comprises three key mechanisms dedicated to RAG:
(i) a vector database layout builds the correlation between embeddings and documents to enable efficient document retrieval for ISP systems, (ii) algorithmic support customized for the ISP-friendly Inverted  File algorithm to improve retrieval performance,
(iii) an in-storage Approximate Nearest Neighbor Search (ANNS) engine to efficiently execute the ANNS kernel. Our evaluation shows that REIS significantly outperforms both (i) a modern CPU-based system for document retrieval and (ii) two state-of-the-art ISP-based ANNS accelerators. We believe and hope that REIS will inspire further research in In-Storage Processing, both in RAG and beyond.

\section*{Acknowledgments}
We sincerely thank Andreas Kosmas Kakolyris for his very significant contributions to the work during and after the rebuttal process. Andreas is a first year PhD student in the SAFARI Research Group who should be a major co-author of the published ISCA 2025 version of this paper. However, due to the policy dictated by the ISCA leadership, which we, as all co-authors, wholeheartedly disagree with and find very problematic, unscientific, and unethical, he was not allowed to be a co-author. We thank the anonymous reviewers of ISCA 2025 for feedback. We thank the SAFARI Research Group members for feedback and the stimulating intellectual environment they provide. We acknowledge the generous gifts from our industrial partners, including Google, Huawei, Intel, and Microsoft. This work is supported in part by the ETH Future Computing Laboratory (EFCL), Huawei ZRC Storage Team, Semiconductor Research Corporation (SRC), AI Chip Center for Emerging Smart Systems (ACCESS), sponsored by InnoHK funding, Hong Kong SAR, and European Union’s Horizon programme for research and innovation [101047160 - BioPIM]. Jisung Park was supported by the National Research Foundation of Korea (RS-2024-00347394, RS-2024-00415602, RS-2024-00459026).

\bibliographystyle{IEEEtranS}
\bibliography{references}

\end{document}